\RequirePackage{fix-cm}
\documentclass[twocolumn]{svjour3}          
\smartqed  
\usepackage{graphicx}%
\usepackage{multirow}%
\usepackage{amsmath,amssymb,amsfonts}%
\usepackage{array}
\usepackage{mathrsfs}%
\usepackage[title]{appendix}%
\usepackage{xcolor}%
\usepackage{textcomp}%
\usepackage{manyfoot}%
\usepackage{booktabs}%
\usepackage{algorithmicx}%
\usepackage{algpseudocode}%
\usepackage{listings}%

\usepackage{array}
\usepackage{stfloats}
\usepackage{url}
\usepackage{verbatim}
\usepackage{graphics} 
\usepackage{algpseudocode}
\usepackage{listings}
\usepackage{xcolor}
\usepackage{float}
\usepackage{lettrine}
\usepackage{subcaption}
\usepackage{arydshln}
\usepackage{natbib}
\usepackage[bookmarks=true,colorlinks=true,urlcolor=blue,citecolor=red]{hyperref}

\usepackage{placeins}   
\usepackage{lipsum}     

\usepackage[ruled,vlined]{algorithm2e}
\usepackage[10pt]{moresize}
\usepackage{threeparttable}

\definecolor{codegreen}{rgb}{0,0.6,0}
\definecolor{codegray}{rgb}{0.5,0.5,0.5}
\definecolor{codepurple}{rgb}{0.58,0,0.82}
\definecolor{backcolour}{rgb}{0.95,0.95,0.92}

\definecolor{cr}{rgb}{1,0,0}
\definecolor{cg}{rgb}{0,0.62,0}
\definecolor{cb}{rgb}{0,0,1}

\lstdefinestyle{mystyle}{
    backgroundcolor=\color{backcolour},   
    commentstyle=\color{codegreen},
    keywordstyle=\color{magenta},
    numberstyle=\tiny\color{codegray},
    stringstyle=\color{codepurple},
    basicstyle=\ttfamily\scriptsize,
    breakatwhitespace=false,         
    breaklines=true,                 
    captionpos=b,                    
    keepspaces=true,                 
    numbers=left,                    
    numbersep=5pt,                  
    showspaces=false,                
    showstringspaces=false,
    inputencoding=utf8,
    extendedchars=false,
    showtabs=false,                  
    tabsize=2
}
\newcommand{\PAR}[1]{\vskip4pt \noindent{\bf #1~}}

\usepackage[normalem]{ulem}
\useunder{\uline}{\ul}{}

\makeatletter
\renewcommand\paragraph{
  \@startsection{paragraph} 
  {4} 
  {\z@} 
  {.5em \@plus1ex \@minus.2ex} 
  {-.5em} 
  {\normalfont\normalsize\bfseries} 
}
\makeatother

\begin{document}
\sloppy

\title{CLIDD: Cross-Layer Independent Deformable Description for Efficient and Discriminative Local Feature Representation

\thanks{This work was supported by the National Natural Science Foundation of China (No. 62403063), and Lejoin Intelligence (Shenzhen) Co., Ltd.}
\thanks{* Corresponding Author: Fenghua He. \\ \email{hefenghua@hit.edu.cn}}
}

\author{Haodi Yao \and
        Fenghua He* \and
        Ning Hao \and
        Yao Su
}


\institute{Haodi Yao \at
                School of Astronautics, Harbin Institute of Technology, China \\
                \email{20B904013@stu.hit.edu.cn}
           \and
           Fenghua He \at
                School of Astronautics, Harbin Institute of Technology, China \\
                \email{hefenghua@hit.edu.cn}
           \and
           Ning Hao \at
                School of Astronautics, Harbin Institute of Technology, China \\
                \email{haoning0082022@163.com}
           \and
           Yao Su \at
                State Key Laboratory of General Artificial Intelligence, Beijing Institute for General Artificial Intelligence (BIGAI), China \\
                \email{yaosu@g.ucla.edu}
}

\date{}

\maketitle

\begin{abstract}
Robust local feature representations are essential for spatial intelligence tasks such as robot navigation and augmented reality. Establishing reliable correspondences requires descriptors that provide both high discriminative power and computational efficiency. To address this, we introduce Cross-Layer Independent Deformable Description (CLIDD), a method that achieves superior discriminability by sampling directly from independent feature hierarchies. 
This approach utilizes learnable offsets to capture fine-grained structural details across scales, without ever constructing a unified dense feature map, eliminating the memory and computation overhead of building a full-scale representation for a sparse sampling process. The per-keypoint per-layer offsets further break interpolation-induced correlation, preserving descriptor discriminability even at high keypoint densities.
To ensure real-time performance, we implement a hardware-aware kernel fusion strategy that maximizes inference throughput. Furthermore, we develop a scalable framework that integrates lightweight architectures with a training protocol leveraging both metric learning and knowledge distillation. This scheme generates a wide spectrum of model variants optimized for diverse deployment constraints. Extensive evaluations demonstrate that our approach achieves superior matching accuracy and exceptional computational efficiency simultaneously. 
Specifically, the ultra-compact variant matches the precision of SuperPoint while utilizing only 0.004M parameters, achieving a 99.7\% reduction in model size. Furthermore, our high-performance configuration outperforms a broad range of current state-of-the-art methods, including high-capacity DINOv2-based frameworks, while exceeding 200 FPS on edge devices. These results demonstrate that CLIDD delivers high-precision local feature matching with minimal computational overhead, providing a robust and scalable solution for real-time spatial intelligence tasks. \footnote{Demo and weights are available at \url{https://github.com/HITCSC/CLIDD}.}
\end{abstract}


\section{Introduction}

Advancements in 3D scene reconstruction and spatial intelligence have revolutionized diverse domains ranging from augmented reality~\citep{NeRF, 3DGS} to autonomous robotics~\citep{NeRFVINS,GaussNav}. Central to these developments is the capability to accurately recover both camera poses~\citep{SfM, HLoc} and spatial geometry from 2D images~\citep{VGGT, DUSt3R, MASt3R}. This capability relies on establishing robust point-wise correspondences across various views, where sparse local feature matching remains the preferred mechanism due to its computational efficiency and structural reliability~\citep{SIFT, ORB, D2-Net, ASLFeat, XFeat}. Specifically, the performance of local feature matching highly depends on the discriminative power of descriptors, which must remain resilient against drastic illumination shifts and large baseline variations~\citep{SuperPoint, DISK}.

\begin{figure*}[htbp]
    \centering
    \begin{minipage}{\textwidth}
        \begin{subfigure}[b]{0.5\textwidth}
            \centering
            \includegraphics[height=0.15\textheight]{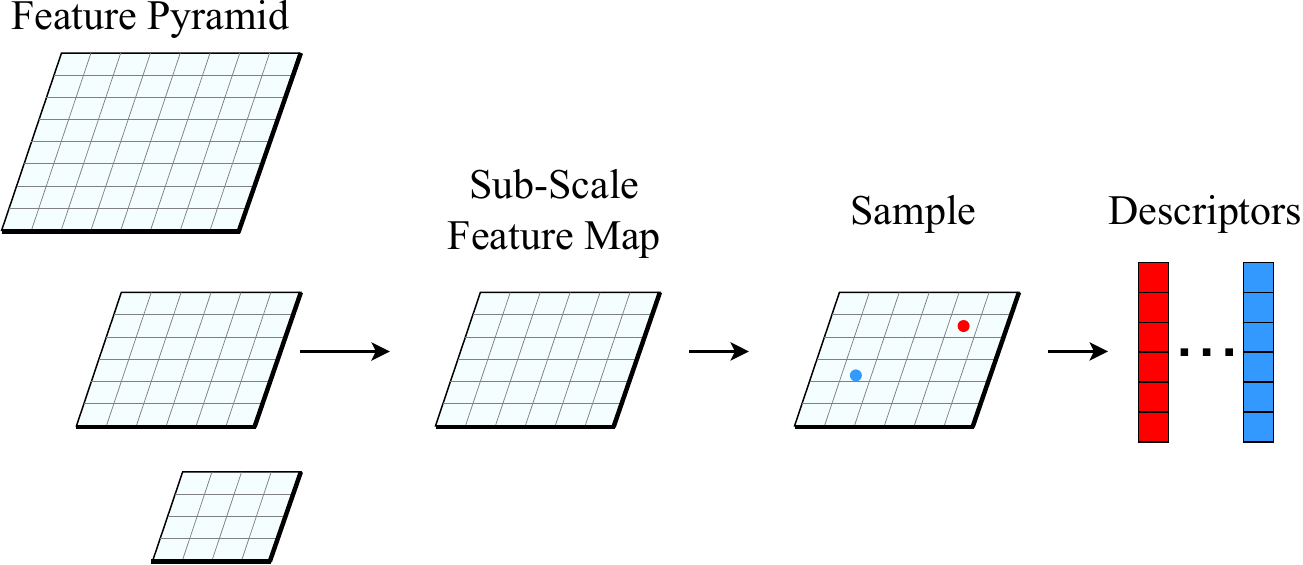}
            \caption{Sub-Scale Vanilla}
            \label{fig:Fig1-a}
        \end{subfigure}
        \hfill
        \begin{subfigure}[b]{0.5\textwidth}
            \centering
            \includegraphics[height=0.15\textheight]{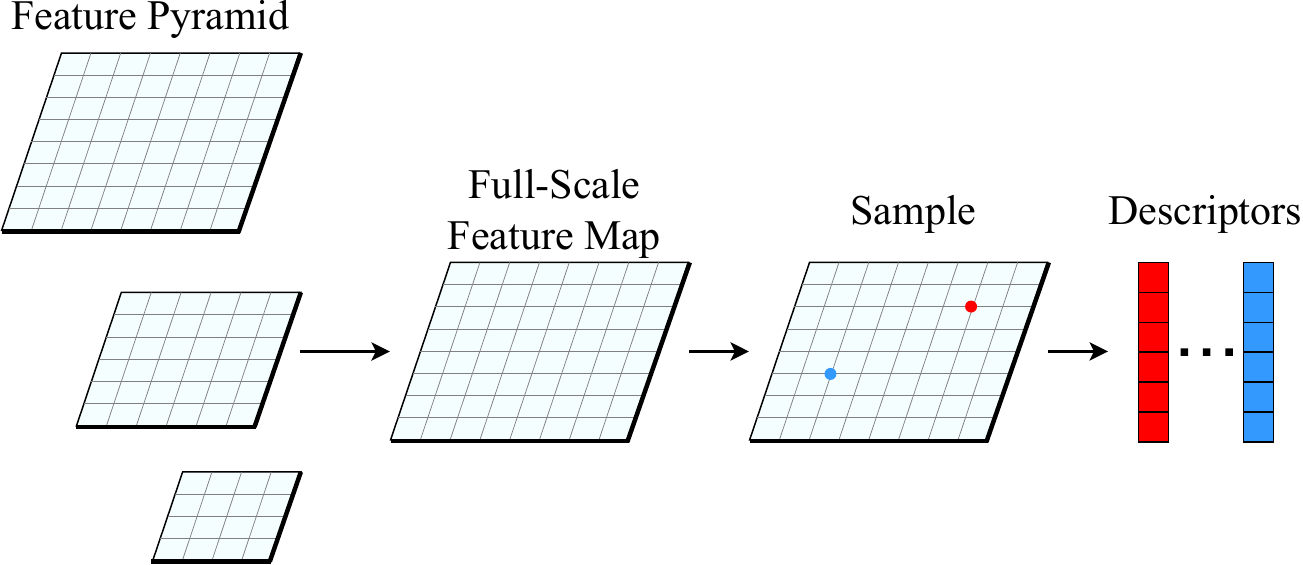}
            \caption{Full-Scale Vanilla}
            \label{fig:Fig1-b}
        \end{subfigure}
        \\[10pt]
        
        \begin{subfigure}[b]{0.5\textwidth}
            \centering
            \includegraphics[height=0.15\textheight]{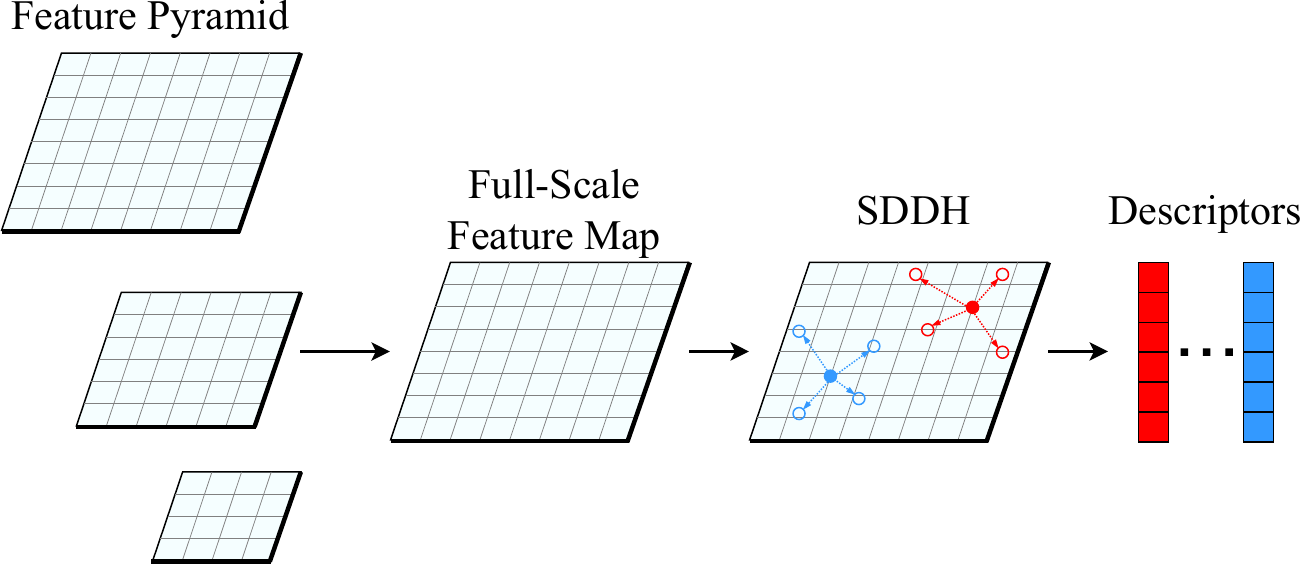}
            \caption{Full-Scale Sparse Deformable Descriptor Head}
            \label{fig:Fig1-c}
        \end{subfigure}
        \hfill
        \begin{subfigure}[b]{0.5\textwidth}
            \centering
            \includegraphics[height=0.15\textheight]{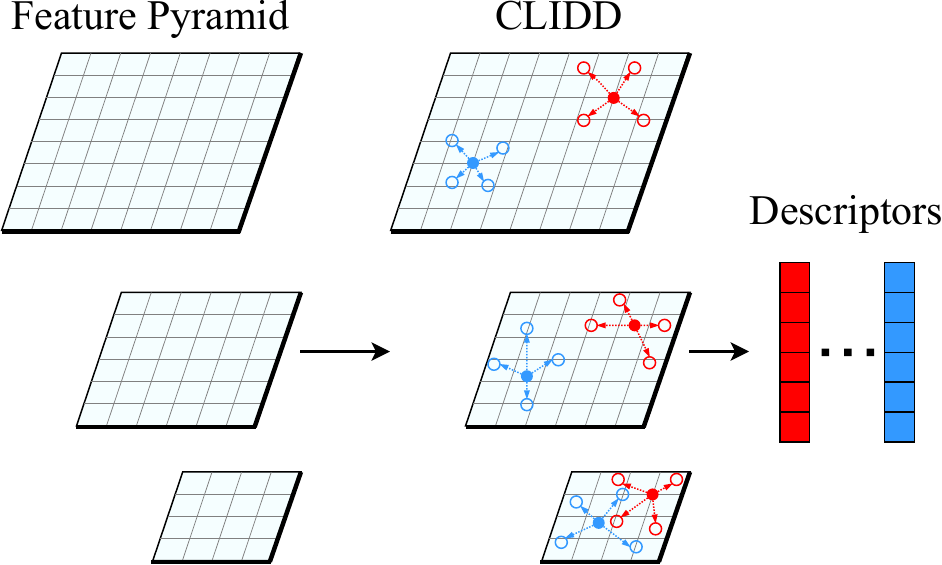}
            \caption{Ours}
            \label{fig:Fig1-d}
        \end{subfigure}
    \end{minipage}
    \caption{\raggedright \textbf{Structural comparison of local descriptor extraction strategies.} (a) Sub-Scale Vanilla aggregates multi-scale features into a single low-resolution dense map for sparse sampling. (b) Full-Scale Vanilla constructs a high-resolution, unified feature field to improve precision at the cost of increased memory overhead. (c) Full-Scale Sparse Deformable Descriptor Head (SDDH)introduces a sparse deformable sampling head on top of the dense, full-scale feature map. (d) Our method employs Cross-Layer Independent Deformable Description (CLIDD) to perform sparse sampling directly from multiple independent feature layers. This strategy generates highly discriminative descriptors while completely bypassing the construction of unified, dense feature maps.}
    \label{fig:Fig1}
\end{figure*}

The design of feature aggregation across network layers shapes both the efficiency and the discriminability of local descriptors. A straightforward approach samples descriptors from a single low-resolution feature map~\citep{XFeat, EdgePoint2}, shown in Fig. \ref{fig:Fig1-a}. Operating on the coarse feature map keeps computational cost low but restricts the spatial resolution available for description. Full-scale vanilla methods instead construct a unified high-resolution feature field to preserve richer detail (Fig. \ref{fig:Fig1-b}), though the memory and computation overhead limits deployment on resource-constrained platforms~\citep{DISK, SiLK, DeDoDe}. The Sparse Deformable Descriptor Head (SDDH) ~\citep{ALIKED} in Fig. \ref{fig:Fig1-c} introduces learnable offsets that decouple sampling from the fixed grid, improving spatial flexibility while still operating on top of a unified dense map. Moreover, all three approaches produce a unified dense feature map whose full-resolution field consumes substantial memory and computation. During the subsequent sampling step, however, only a sparse subset of spatial positions is ever accessed. Most of the stored feature vectors are therefore left unused, creating an inherent mismatch between representation cost and utility.

Fusing multiple feature scales into a unified map creates a representational problem that extends beyond computational cost. Shallow network layers capture fine-grained local structures, while deep layers integrate broader contextual and semantic information through their wider receptive fields. When these distinct feature hierarchies are concatenated into a single representation, their layer-specific characteristics become entangled and indistinguishable. The subsequent sampler can no longer determine which features originate from which scale, preventing each layer from developing its own specialized offset strategy tailored to its discriminative feature properties.

A further limitation arises from local correlation introduced by spatial interpolation. When the unified feature map resolution is coarser than the original image, extracting descriptors at precise pixel coordinates requires interpolation from neighboring grid positions. Nearby keypoints falling within the same interpolation neighborhood draw from an overlapping set of feature vectors, causing their descriptors to share substantial local structure that erodes pairwise discriminability. Restoring features to full image resolution avoids this problem but entails prohibitive computational cost.

\begin{figure*}
    \centering
    \includegraphics[width=\textwidth]{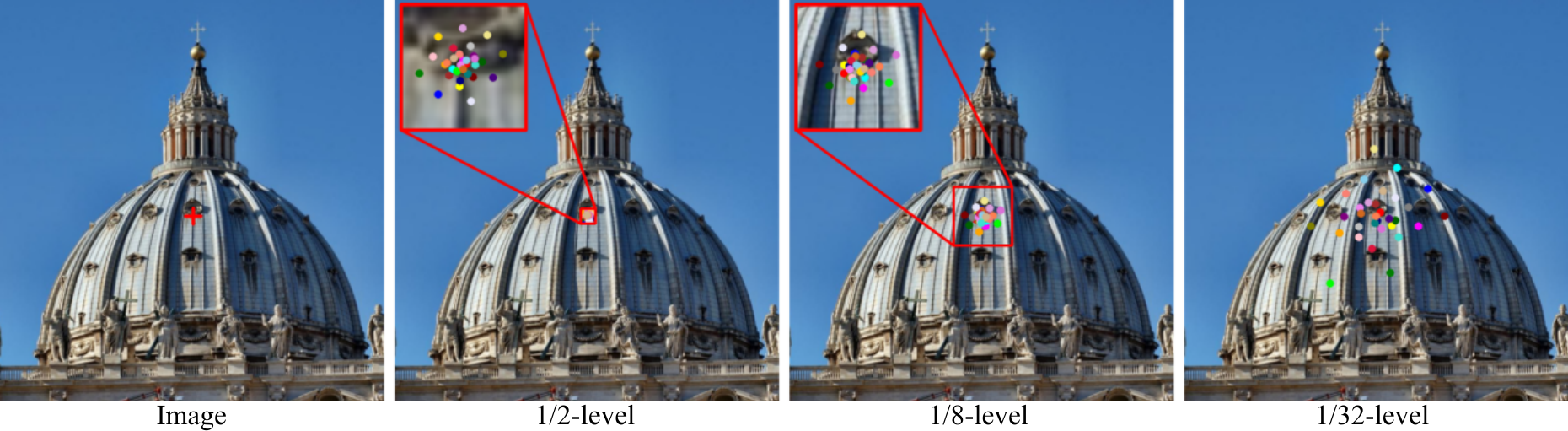}
    \caption{\raggedright \textbf{Deformable sampling offsets of our U128 model on three independent feature maps at 1/2-level, 1/8-level, and 1/32-level.} All offsets are projected and visualized at the original image scale for comparison. }
    \label{fig:front_cover_defrom}
\end{figure*}

These three limitations motivate a description strategy that samples directly from independent feature layers without constructing a unified map. In this work, we present Cross-Layer Independent Deformable Description (CLIDD), a method that generates discriminative descriptors directly from multi-scale hierarchies, as shown in Fig. \ref{fig:Fig1-d}. CLIDD operates on independent feature layers through two coordinated components. A Cross-Layer Predictor leverages information across all scales to produce sampling offsets tailored to each layer, allowing each scale to benefit from cross-layer context. A Layer-Independent Sampler then extracts the discriminative features native to each resolution level, as shown in Fig.~\ref{fig:front_cover_defrom}. This architecture enables CLIDD to break local correlation and strengthen the discriminability of descriptors. To further maximize efficiency, we implement a kernel fusion strategy that integrates sampling and aggregation into a single operation, eliminating the memory and computation overhead of intermediate dense representations. We integrate CLIDD into a scalable framework that couples lightweight architectures with a training protocol combining metric learning and knowledge distillation. Specifically, we employ DualSoftmax loss for robust description alongside UnfoldSoftmax loss for lightweight detection. We further employ Orthogonal-Procrustes loss for the most compact variants, ensuring high precision despite a minimal parameter count.

\begin{figure}
    \centering
    \includegraphics[width=\columnwidth]{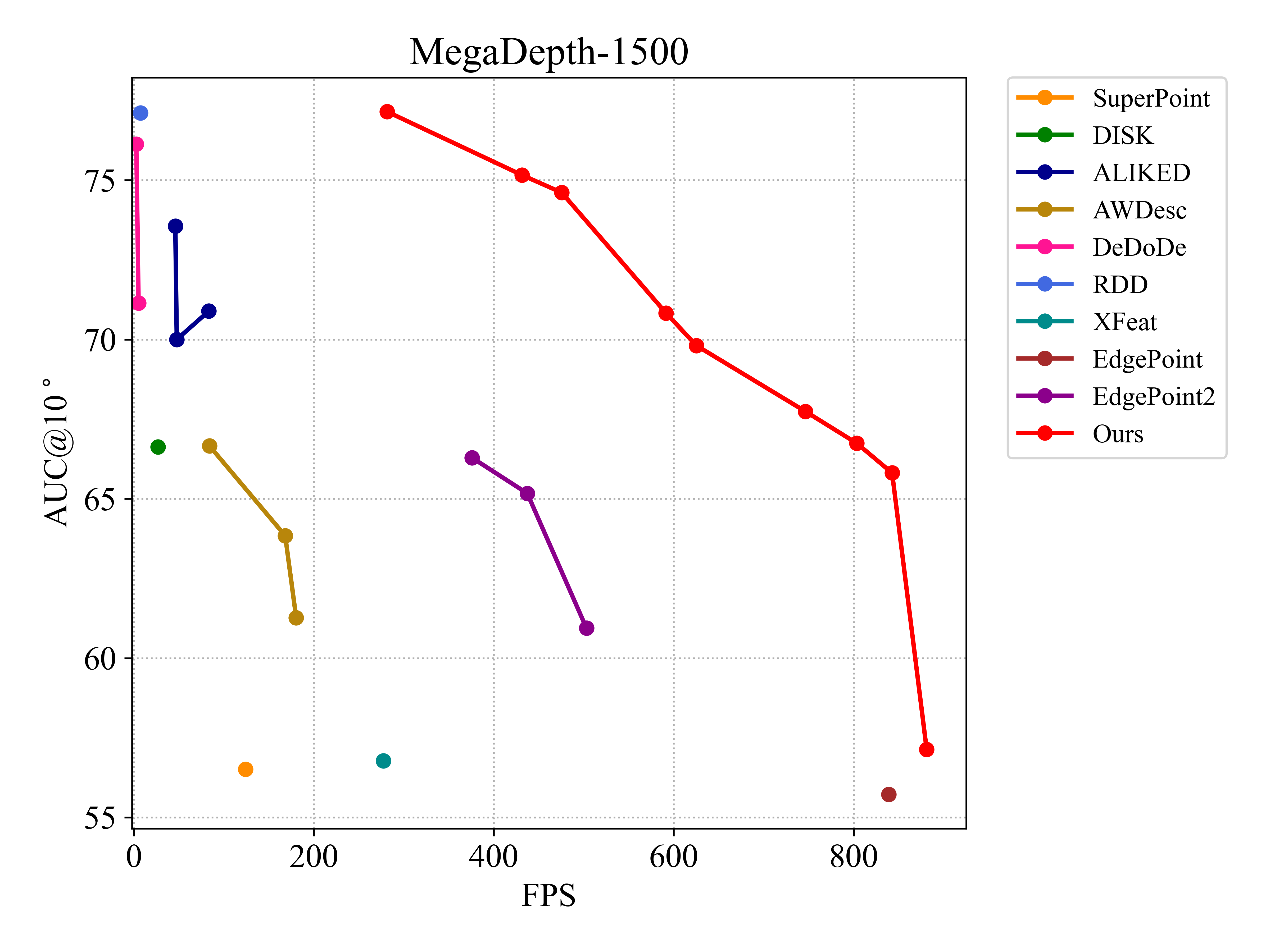}
    \caption{\raggedright \textbf{Precision-efficiency comparison on resource-constrained devices.} This plot illustrates the relationship between relative pose estimation accuracy (AUC@10$^\circ$) on the MegaDepth-1500 benchmark and inference speed (FPS) measured on an edge platform. Our CLIDD-based models consistently occupy the upper-right quadrant, outperforming their corresponding state-of-the-art baselines in both matching precision and processing speed across model scales.}
    \label{fig:fps_vs_megadepth}
\end{figure}

The performance of CLIDD is validated through comprehensive experiments across benchmarks for homography estimation, relative pose estimation, and large-scale visual localization. As illustrated in Fig.~\ref{fig:fps_vs_megadepth}, our models consistently outperform their corresponding state-of-the-art baselines by delivering superior matching precision and exceptional processing speeds simultaneously. Specifically, our ultra-compact variant matches the precision of SuperPoint~\citep{SuperPoint} with only 0.004M parameters, achieving a 99.7\% reduction in model size. Furthermore, the high-performance configuration outperforms state-of-the-art methods, including high-capacity DINOv2-based frameworks~\citep{DeDoDe}, while exceeding 200 FPS on resource-limited edge devices. Extensive ablation studies and high-density sampling evaluations confirm that our design achieves gains in both accuracy and computational efficiency.

The main contributions are summarized as follows:

\begin{itemize}
    \item We propose Cross-Layer Independent Deformable Description, which produces sparse descriptors directly from independent multi-scale layers via a Cross-Layer Predictor and a Layer-Independent Sampler. This design bypasses the construction of a unified dense feature map, strengthening descriptor discriminability while eliminating the associated memory and computation overhead. A kernel fusion strategy further integrates sampling and aggregation into a single operation for real-time inference.

    \item We introduce a scalable framework that integrates lightweight architectures with a composite training protocol of metric learning and knowledge distillation, yielding diverse model variants optimized for various hardware constraints.

    \item Extensive evaluations across multiple benchmarks demonstrate that our strategy achieves superior matching precision and efficiency. The proposed descriptors deliver leading matching accuracy together with substantial speed advantages over existing high-capacity models, even under challenging high-density sampling conditions.
\end{itemize}

\section{Related Work}

\subsection{Keypoint Description}

The evolution of local descriptors has been largely shaped by the trade-offs between spatial resolution and computational demand. Early learning-based frameworks ~\citep{SuperPoint,D2-Net,KP2D} established the standard of extracting descriptors via interpolation on low-resolution feature maps. Efficiency-oriented models ~\citep{XFeat,EdgePoint2} have since optimized this strategy to bypass high-resolution processing bottlenecks. However, relying on a coarse grid often restricts the ability to distinguish fine-grained image structures, as the under-sampled feature volumes provide limited information for generating highly discriminative representations.

To maximize matching precision, high-performance methods such as DISK and ALIKE~\citep{DISK,ALIKE} utilize dense, high-resolution feature maps. While this approach improves results, it introduces significant memory and processing overhead. Recent advancements such as AWDesc~\citep{AWDesc} integrate Transformer-based components to strengthen representation quality. Specialized architectures such as DeDoDe~\citep{DeDoDe} further separate detection and description, with the DeDoDe-G variant scaling this architecture using the DINOv2~\citep{DINOv2} foundation model. Despite these performance gains, these frameworks still rely on generating a unified, full-resolution description field to maintain their accuracy.

Sparse deformable sampling provides a strategic alternative for improving efficiency. ALIKED~\citep{ALIKED} introduces a dedicated sparse head that avoids the repetitive convolutions required by traditional methods on dense maps. This mechanism utilizes learnable offsets to gather information more flexibly than a fixed coordinate grid. Nevertheless, existing deformable approaches still aggregate multi-scale features into a single dense representation before sampling occurs. This intermediate fusion not only constitutes a computational bottleneck but also entangles layer-specific characteristics. Once features from different depths are concatenated into one volume, the sampler can no longer distinguish shallow localized contributions from deep contextual information, preventing each scale from developing its own specialized offset strategy.

Across these paradigms, descriptor extraction ultimately depends on a unified dense feature map that is constructed before any sampling takes place. This feature map typically has a channel count equal to the final descriptor dimension. The full spatial extent multiplied by this high dimensionality yields a substantial memory footprint, yet the map must be stored in its entirety even though only a sparse set of its positions is ever queried. Moreover, if the fused map is smaller than the original image in resolution, interpolation at keypoint locations further causes adjacent descriptors to draw from overlapping feature regions, weakening their pairwise discriminability.

In this work, we introduce a description method that addresses these limitations by combining cross-layer deformable sampling with direct extraction from multiple feature scales. By decoupling sampling positions across the hierarchy, our approach generates highly expressive descriptors without the need to aggregate features into a single volume. This strategy bypasses the construction of unified feature maps entirely, allowing our framework to preserve sharp structural details while maintaining a minimal memory footprint and high execution speed.

\subsection{Deformable Operation}

Deformable operations have become essential for modeling spatial transformations and geometric variations. Deformable Convolutional Networks (DCN)~\citep{DCNv1} first introduced learnable offsets to adapt kernel receptive fields to complex object structures. Subsequent iterations ~\citep{DCNv2, DCNv4} advanced these mechanisms by incorporating modulation components and hardware-efficient implementations. Deformable Attention~\citep{DeformableAttention} later extended these principles into the Transformer domain, enabling content-aware adaptive attention mechanisms. To overcome the inefficiency of naive implementations, these methods incorporate kernel fusion techniques. These developments provide the technical foundation for adaptive feature extraction in modern vision systems.

In the context of local feature learning, ASLFeat~\citep{ASLFeat} introduced deformable convolutions within backbone layers to enhance the geometric invariance of keypoint representations. This architecture ensures that descriptors remain stable across significant image transformations. Building upon this, ALIKED~\citep{ALIKED} integrates deformable modules within its feature extractor alongside a sparse sampling strategy. While this sampling head allows for more flexible feature gathering at specific interest points, it remains dependent on a unified feature map generated through resource-intensive aggregation.

More recently, RDD~\citep{RDD} employs Deformable Attention~\citep{DeformableAttention} within the descriptor extraction network to capture geometrically invariant features. Its deformable attention modules operate across multiple feature scales simultaneously, enabling cross-layer interaction that integrates global contextual information. However, this cross-scale deformable computation is performed through several consecutive attention blocks, each operating on dense feature grids. Stacking multiple such blocks compounds the computational cost despite hardware acceleration.

In contrast, our approach utilizes a streamlined backbone based on standard convolutions to prioritize inference throughput. We shift the focus from backbone deformation to the description stage through Cross-Layer Independent Deformable sampling. Unlike previous methods that apply deformable computation on dense feature grids, our framework restricts deformable sampling to the sparse set of detected keypoints, directly indexing into multiple independent feature maps. By simultaneously performing offset prediction and feature extraction across different hierarchies, we achieve high matching precision while maintaining exceptional computational efficiency.

\section{Cross-Layer Independent Deformable Description}

\begin{figure*}[htbp]
    \centering
    \includegraphics[width=\textwidth]{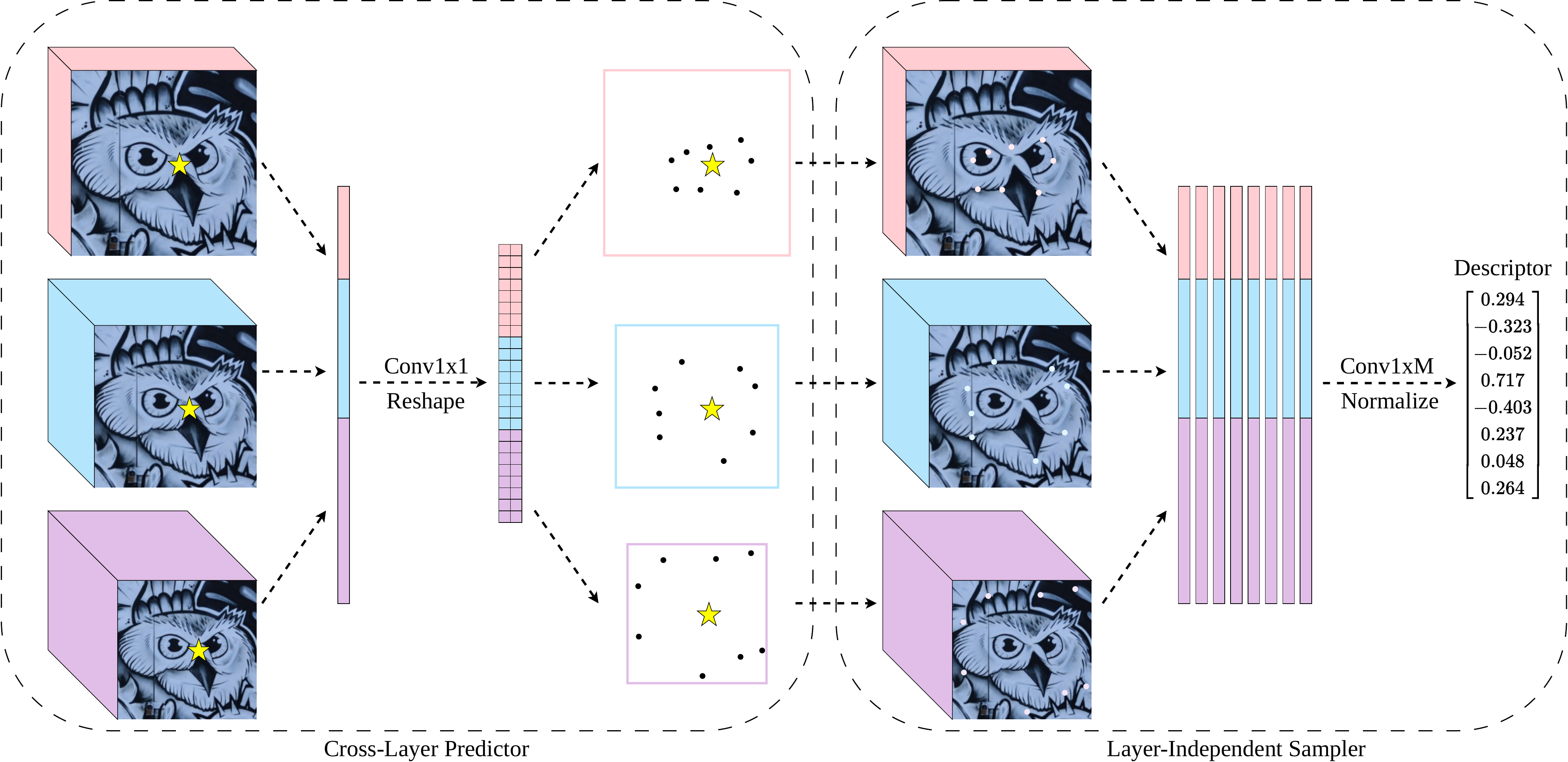}
    \caption{\raggedright \textbf{Mechanism of Cross-Layer Independent Deformable Description.} This framework facilitates sparse feature extraction across multiple hierarchies without relying on a unified dense feature map. The Cross-Layer Predictor utilizes spatially aligned embeddings to determine level-specific sampling offsets for each feature resolution. Guided by these offsets, the Layer-Independent Sampler performs decoupled feature retrieval from each layer, where $M$ denotes the number of deformable sampling points per scale. This mechanism allows the model to capture both fine-grained geometry and semantic context simultaneously. The independently sampled features are then aggregated to produce a discriminative descriptor. While illustrated with three scales, this design is inherently scalable and generalizes to standard deformable sampling in single-layer configurations.}
    \label{fig:clidd}
\end{figure*}

In this section, we introduce the Cross-Layer Independent Deformable Description (CLIDD) method, as illustrated in Fig. \ref{fig:clidd}, which generates highly discriminative descriptors through efficient cross-layer sparse sampling. We first describe the Cross-Layer Predictor, which coordinates precise sampling offsets by aggregating information across multiple feature scales. This is followed by the Layer-Independent Sampler, which extracts features from separate hierarchies using decoupled sampling points to capture multi-scale characteristics. Finally, we detail the kernel fusion techniques employed to optimize memory access patterns and ensure high-throughput inference for the entire module.

\subsection{Cross-Layer Predictor}

The CL-Predictor determines offsets by operating directly on feature vectors rather than processing dense, unified feature maps. For each keypoint, the predictor extracts a single feature vector from every available layer using normalized coordinates. These individual vectors are concatenated into a cross-layer embedding that represents the structural context across the hierarchy. This design bypasses the need for resizing or stacking large feature maps and thereby significantly reduces computational and memory costs. By operating on these feature vectors directly, the predictor removes the heavy overhead typically associated with dense grid calculations.

The concatenated embedding is processed through a point-wise convolution layer to generate a comprehensive set of sampling offsets. These offsets are partitioned into groups, so that each feature layer receives its own set of $M$ unique coordinates. Shallow layers receive compact local offsets aligned to their narrow receptive fields, while deep layers receive broader offsets that cover their wider contextual scope. This coordinated prediction allows the model to leverage information from all scales to determine the most relevant sampling locations for each specific resolution. This strategy proves more effective than independent per-layer prediction while remaining significantly lighter than dense feature fusion. The CL-Predictor thus provides essential coordinate biases while maintaining a minimal computational footprint.

\subsection{Layer-Independent Sampler}

The LI-Sampler utilizes the specific offsets generated by the predictor to execute independent feature extraction across different hierarchies. Most traditional frameworks aggregate all feature levels into a single dense representation before any sampling occurs. Our approach avoids this fusion step entirely, thereby preserving the unique information present at each resolution level. By utilizing independent sampling offsets for each layer, the model gains the flexibility to capture diverse structural characteristics across the feature hierarchy.

Following the extraction, features from all layers are integrated into a final descriptor through a unified aggregation process. This method produces a more diverse representation of the local region compared to samplers relying on fixed or shared geometries. This decoupling of sampling positions allows the model to gather unique information across the hierarchy while keeping the overall system highly efficient. Moreover, each keypoint receives independent deformable offsets for every layer. As a result, adjacent keypoints are likely assigned distinct sampling biases at each scale, breaking the local correlation that bilinear interpolation introduces between neighboring descriptors.

\subsection{Kernel Fusion and Efficient Implementation}

Standard implementations of multi-scale description heads often encounter hardware bottlenecks due to excessive global memory operations. In a typical setup, sampled features from every layer are stored as intermediate tensors before the final aggregation. When processing a large number of feature points or using high-dimensional feature layers, this approach creates substantial memory pressure. The frequent reading and writing of these intermediate results to global memory leads to severe latency penalties and limits the throughput of the model.

To overcome these architectural constraints, we implement a custom fused kernel that integrates feature sampling and aggregation into a single hardware operation. The kernel processes keypoints in small subsets, sampling the independent coordinates and performing partial aggregation directly within the fast on-chip SRAM. This design effectively decomposes the aggregation weight matrix into layer-wise components, allowing the contribution of each layer to be computed locally. By eliminating the need to store large intermediate tensors in global memory, the fused implementation minimizes the peak memory footprint and significantly accelerates inference speed across various computational platforms. The kernel further reorders tensors to a channels-last memory layout, placing the feature vector of each spatial position contiguously in memory. During bilinear interpolation, this layout allows the four corner feature vectors of each enclosing grid cell to be fetched in a single coalesced read, avoiding strided memory accesses across the channel dimension. Furthermore, the kernel processes multiple sampling points jointly within a single launch. This joint processing increases the reuse ratio of weight parameters and exposes a larger volume of regularly structured computation, allowing the GPU Tensor Cores to operate closer to peak throughput.

\section{Model Architecture and Training Scheme}

This section details the architectural design and the training strategy of our framework. We first introduce a scalable architecture that integrates our core description method with optimized components to simultaneously achieve high-fidelity representation and fast inference. The training process then employs a composite objective function to optimize descriptor discriminability and geometric accuracy through metric learning and distillation. Finally, we provide the implementation details and training protocols that ensure robust performance across all model configurations.

\begin{figure*}[htbp]
    \centering
    \includegraphics[width=\textwidth]{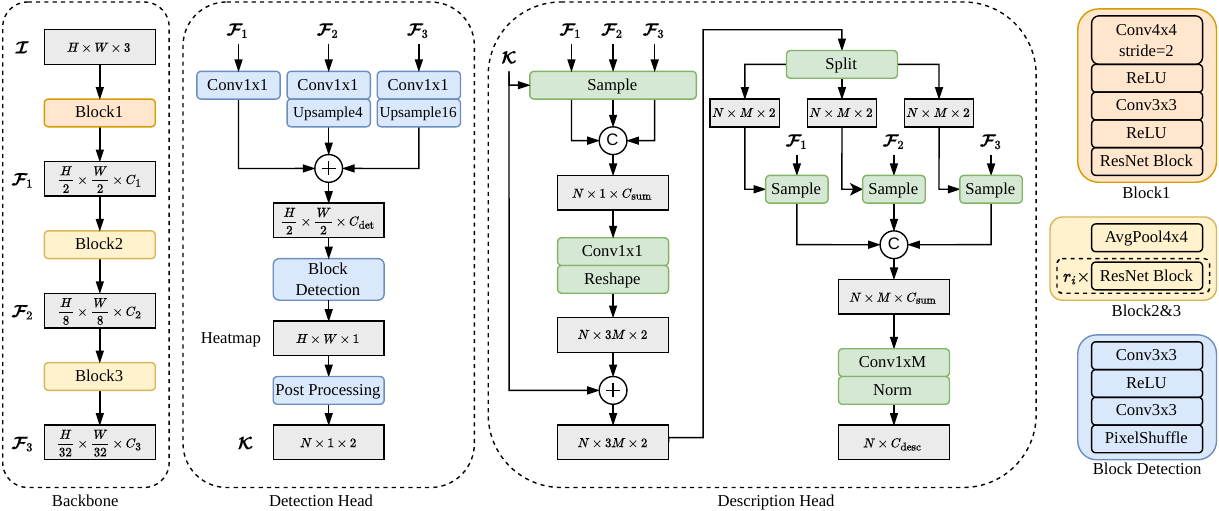}
    \caption{\raggedright \textbf{Model Architecture.} Our lightweight framework generates multi-scale representations using standard convolutional layers and ResNet blocks. The notation $C_i$ for $i \in \{1,2,3\}$ specifies the output channel counts for each feature block, while $C_{\mathrm{det}}$ and $C_{\mathrm{desc}}$ denote the dimensions for keypoint detection and final description. Within the description head, the total dimension of concatenated sampled features is defined as $C_{\mathrm{sum}} = C_1 + C_2 + C_3$. Symbols $\textcircled{\scalebox{0.8}{+}}$ and $\textcircled{\scalebox{0.7}{C}}$ indicate element-wise addition and concatenation operations, respectively. The term $r_i$ for $i \in \{2,3\}$ determines the number of ResNet blocks within each specific stage. Following extraction, a post-processing module transforms the dense detection heatmap into sparse keypoints. By utilizing our Cross-Layer Independent Deformable Description method, the architecture produces high-quality descriptors without generating dense, high-resolution feature maps, thereby simultaneously enhancing both descriptive precision and computational efficiency.}
    \label{fig:model-architecture}
\end{figure*}

\subsection{Lightweight Model Architecture}

We implement the proposed Cross-Layer Independent Deformable Description method as the core descriptor within a highly scalable network, as shown in Fig. \ref{fig:model-architecture}. This design integrates a lightweight backbone and a streamlined detection head to achieve superior representation quality without compromising computational efficiency. The resulting architecture ensures robust performance across varying hardware constraints while maintaining the high throughput required for real-time deployment.

\subsubsection{Feature Backbone}

Our framework employs a standardized encoder to generate a multi-scale feature hierarchy at 1/2, 1/8, and 1/32 downsampling ratios. The process begins with a $4\times4$ convolution with a stride of 2 to reduce spatial resolution early and minimize initial computational overhead. At the 1/2 resolution stage, a single ResNet block~\citep{ResNet} captures essential local details while maintaining a minimal memory footprint. Subsequent deeper stages utilize average pooling to reach 1/8 and 1/32 resolutions, allowing the capture of broader global context without excessive complexity.

We scale the number of residual blocks and channel widths at lower resolutions to balance semantic depth with execution speed. All configurations utilize ReLU activation functions to ensure the highest possible inference throughput on diverse hardware. The design deliberately keeps high-resolution stages shallow to avoid memory bottlenecks associated with processing large feature maps. This optimized multi-scale structure provides the necessary foundation for the subsequent detection and description tasks.

\subsubsection{Keypoint Detection Head}

The detection head is standardized across all model tiers to operate at 1/2 resolution for an optimal balance between localization accuracy and efficiency. This module applies $1\times1$ convolutions to the feature pyramid to compress channel dimensions before merging them through element-wise addition. This fusion mechanism efficiently integrates multi-scale context without increasing the total parameter count.

Following feature refinement through additional convolutional layers, a PixelShuffle operation restores the output to the original image dimensions. This technique eliminates upsampling artifacts and provides high-fidelity localization while avoiding the high computational cost of transposed convolutions. The unified head design ensures that even the largest models maintain real-time performance on resource-constrained platforms.

\subsubsection{Cross-Layer Independent Deformable Description Head}

The description head utilizes the multi-scale hierarchy to construct final descriptors of dimension $C_{\mathrm{desc}}$. At each detected keypoint, the cross-layer predictor identifies optimal sampling coordinates across the 1/2, 1/8, and 1/32 feature levels. This mechanism captures both local structural details and broad semantic context without requiring resource-intensive upsampling layers. By operating directly on each feature map at its native resolution, this design bypasses the computational and memory bottlenecks of unified feature aggregation. The layer-independent sampling strategy further assigns independent offsets across the hierarchy, preserving descriptor discriminability even when keypoints lie close together.

\subsection{Loss Functions}

Our framework is trained using image pairs with known geometric relationships to establish reliable ground-truth correspondences. The total training objective combines metric learning and knowledge distillation through three specific components: the DualSoftmax loss, the Orthogonal-Procrustes loss, and the UnfoldSoftmax loss. Together, these functions optimize the feature discriminability and localization precision required for efficient keypoint extraction.

\subsubsection{DualSoftmax Loss}

The DualSoftmax loss provides the primary supervision for learning discriminative descriptors using image pairs $\boldsymbol{\mathcal{I}}_A$ and $\boldsymbol{\mathcal{I}}_B$ with known geometric relationships. For each training sample, we first extract 1024 keypoints from $\boldsymbol{\mathcal{I}}_A$ using the ALIKED-N32\citep{ALIKED} method. The network then generates the corresponding descriptor matrix $\boldsymbol{D}_A \in \mathbb{R}^{N \times C_\mathrm{desc}}$ for these locations, where $N$ denotes the total number of extracted descriptors. Using the known transformation between the pair, we warp the keypoint coordinates to $\boldsymbol{\mathcal{I}}_B$ and extract the paired descriptor matrix $\boldsymbol{D}_B \in \mathbb{R}^{N \times C_\mathrm{desc}}$. Since some points may move outside the visible bounds of the second image, we define a binary visibility mask $\boldsymbol{m} \in \{0,1\}^N$ to identify valid correspondences.

To evaluate the similarity between the extracted feature sets, we compute the cosine distance matrix $\boldsymbol{S} \in \mathbb{R}^{N \times N}$ as $\boldsymbol{S} = \boldsymbol{D}_A \boldsymbol{D}_B^\mathrm{T}$. The matching probability matrix $\boldsymbol{P}$ is then derived through a dual-softmax operation. This operation enforces mutual exclusivity by independently normalizing the similarity scores across both rows and columns:
\begin{equation}
\begin{aligned}
\boldsymbol{P} = \operatorname{softmax}_r(\frac{\boldsymbol{S}}{T}) \cdot \operatorname{softmax}_c(\frac{\boldsymbol{S}}{T}),
\end{aligned}
\end{equation}
where $\operatorname{softmax}_r$ and $\operatorname{softmax}_c$ denote the softmax functions applied along the rows and columns of the matrix respectively. A fixed temperature constant $T = 0.05$ is utilized to concentrate the resulting probability distribution.

The DualSoftmax loss is formulated as the negative log-likelihood of the ground-truth matches:
\begin{equation}
\begin{aligned}
L_{\mathrm{DS}} = - \frac{1}{\boldsymbol{1}^{\mathrm{T}}\boldsymbol{m}} \boldsymbol{m}^\mathrm{T}\operatorname{log} \operatorname{diag}(\boldsymbol{P}).
\end{aligned}
\end{equation}
where $\operatorname{diag}(\cdot)$ extracts the diagonal elements, and $\operatorname{log}(\cdot)$ denotes the element-wise logarithm. This objective constrains the embedding space to minimize the distance between corresponding features while effectively separating non-matching pairs. By optimizing this probability manifold, the network learns to produce highly unique feature representations. This process effectively maintains consistency between corresponding points while maximizing the distance between distinct regions.

\subsubsection{Orthogonal Procrustes Loss}

Super-lightweight model configurations often require external guidance to realize their full discriminative potential. To address this, we employ knowledge distillation by supervising these efficient variants with a teacher model. We select ALIKED-N32~\citep{ALIKED} as the teacher architecture, which possesses a fixed descriptor dimension $C'_\mathrm{teacher}$ of 128. During training, we utilize the keypoint locations from the DualSoftmax phase to extract student descriptors $\boldsymbol{\hat{D}}_A \in \mathbb{R}^{(B\cdot N) \times C_\mathrm{desc}}$ and teacher descriptors $\boldsymbol{\hat{D}}_\mathrm{teacher} \in \mathbb{R}^{(B\cdot N) \times C'_\mathrm{teacher}}$ across the batch. Because our student descriptors possess a lower dimensionality than those of the teacher ($C_\mathrm{desc} < C'_\mathrm{teacher}$), we adopt the Orthogonal-Procrustes loss framework proposed in~\citep{EdgePoint2} for cross-dimensional distillation. A standard distillation target focuses on the internal distances within the embedding space, defined as:
\begin{equation}
\begin{aligned}
    \| \boldsymbol{\hat{D}}_A \boldsymbol{\hat{D}}_A^\mathrm{T} - \boldsymbol{\hat{D}}_\mathrm{teacher} {\boldsymbol{\hat{D}}_\mathrm{teacher}}^\mathrm{T} \|_\mathrm{F}^2. \\
\end{aligned}
\label{DistillationTargetNaive}
\end{equation}

However, this formulation lacks direct geometric alignment. To distill the geometric structure more effectively, we follow~\citep{EdgePoint2} and apply a Low-Rank Approximation (LRA) to compress the rank of $\boldsymbol{\hat{D}}_\mathrm{teacher} {\boldsymbol{\hat{D}}_\mathrm{teacher}}^\mathrm{T}$ to $C_\mathrm{desc}$. We compute the Singular Value Decomposition (SVD) of the teacher descriptors as $\boldsymbol{\hat{D}}_\mathrm{teacher} = \boldsymbol{U} \boldsymbol{\Sigma} \boldsymbol{V}^\mathrm{T}$. We then construct the compressed descriptors using the leading $C_\mathrm{desc}$ singular vectors and values, denoted as $\boldsymbol{\hat{D}}_l = \boldsymbol{U}_C \boldsymbol{\Sigma}_C$. Since the number of points $N$ consistently exceeds the feature dimension $C_\mathrm{desc}$, this compression is inherently lossy. To accelerate training, we perform this LRA per batch rather than per individual image. Crucially, as lossy compression does not preserve unit length, we normalize $\boldsymbol{\hat{D}}_l$ along the feature dimension so that its row vectors lie on the unit hypersphere, denoted as $\boldsymbol{\hat{D}}_n$. The distillation objective is subsequently updated as:
\begin{equation}
\begin{aligned}
    \| \boldsymbol{\hat{D}}_A \boldsymbol{\hat{D}}_A^\mathrm{T} - \boldsymbol{\hat{D}}_n {\boldsymbol{\hat{D}}_n}^\mathrm{T} \|_\mathrm{F}^2. \\
\end{aligned}
\label{DistillationTargetLRA}
\end{equation}

The optimal solution seeks a transformation $\boldsymbol{\hat{D}}_A^*=\boldsymbol{\hat{D}}_n\boldsymbol{X}$ where $\boldsymbol{X}$ belongs to the Orthogonal group $\mathbb{O}(C_\mathrm{desc},\mathbb{R})$. We identify the optimal orthogonal matrix $\boldsymbol{\Omega}$ by solving the Orthogonal-Procrustes Problem (OPP)~\citep{OPP}:
\begin{equation}
\begin{aligned}
    \boldsymbol{\Omega} &= \operatorname*{arg\,min}_{\boldsymbol{X} \in \mathbb{O}(C_\mathrm{desc},\mathbb{R})} \ \|\boldsymbol{\hat{D}}_\mathrm{n} \boldsymbol{X} - \boldsymbol{\hat{D}}_A\|_\mathrm{F}^2 \\
             &= \operatorname*{arg\,max}_{\boldsymbol{X} \in \mathbb{O}(C_\mathrm{desc},\mathbb{R})} \ \mathrm{tr}(\boldsymbol{\hat{D}}_A^\mathrm{T} \boldsymbol{\hat{D}}_\mathrm{n} \boldsymbol{X}) \\
             &= \boldsymbol{V}_p \boldsymbol{U}_p^\mathrm{T},
\end{aligned}
\end{equation}
where $\boldsymbol{U}_p \boldsymbol{\Sigma}_p \boldsymbol{V}_p^\mathrm{T}$ is the SVD of the correlation matrix $\boldsymbol{\hat{D}}_A^\mathrm{T} \boldsymbol{\hat{D}}_\mathrm{n}$. The final Orthogonal-Procrustes loss is formulated as:
\begin{equation}
\begin{aligned}
L_{\mathrm{OP}} &= \| \boldsymbol{1} - \operatorname{diag}(\boldsymbol{\hat{D}}_n\boldsymbol{\Omega}\boldsymbol{\hat{D}}_A^\mathrm{T})\|_\mathrm{2}^2.
\end{aligned}
\end{equation}
Through this scheme, our compact models effectively inherit the robust representational properties encoded by high-capacity teacher architectures. While this batch-wise formulation differs from the lossless compression in~\citep{EdgePoint2}, executing SVD operations per batch significantly accelerates training. The integration of DualSoftmax supervision provides robust discriminative guidance, ensuring that the model maintains high matching accuracy despite the lossy nature of the compressed distillation target.

\subsubsection{UnfoldSoftmax Loss}

The detection training aims to achieve precise keypoint localization while ensuring that inference remains computationally efficient. We supervise the raw output logits directly to enable the network to produce discriminative response maps without additional activation layers. This approach ensures reliable keypoint localization while maintaining a lean execution profile for deployment.

We implement the UnfoldSoftmax loss $L_{\mathrm{US}}$ following the distillation framework proposed in~\citep{EdgePoint2}. This methodology leverages the ALIKED-N32 model to provide high-quality detection labels on image $\boldsymbol{\mathcal{I}}_A$ for supervision. The training process effectively treats local neighborhood patches as competitive classification tasks to refine keypoint precision. We utilize efficient convolutional operations to compute this loss, which ensures numerical stability without increasing the memory footprint. This strategy allows the lightweight detector to inherit the robust localization capabilities of the teacher model while remaining optimized for real-time performance.

\subsection{Implementation Details}

\begin{table*}[t]
\centering
\caption{\textbf{Network configurations.} We denote the models as A/N/T/S/M/L/G/E/U to indicate the model size.}
\begin{tabular}{lcccccccccccc}
\hline
\multicolumn{1}{c}{\multirow{2}{*}{\textbf{Model}}} & \multicolumn{8}{c}{\textbf{Network Configuration}}                                     & \multicolumn{4}{c}{\textbf{Params (MP)}} \\
\multicolumn{1}{c}{}                                & $C_1$ & $C_2$ & $C_3$ & $r_2$ & $r_3$ & $C_{\mathrm{det}}$ & $M$ & $C_{\mathrm{desc}}$ & Backbone   & Detect    & Desc   & Total  \\ \hline
A48(Atom)                                           & 4     & 4     & 4     & 1     & 1     & 4                  & 4   & 48                  & 0.00123    & 0.00036   & 0.003  & 0.004  \\
N64(Nano)                                           & 8     & 8     & 8     & 1     & 1     & 8                  & 8   & 64                  & 0.00448    & 0.00109   & 0.014  & 0.019  \\
T64(Tiny)                                           & 8     & 16    & 24    & 1     & 1     & 8                  & 8   & 64                  & 0.015      & 0.00128   & 0.027  & 0.043  \\
S64(Small)                                          & 8     & 24    & 32    & 1     & 1     & 8                  & 16  & 64                  & 0.026      & 0.00141   & 0.072  & 0.100  \\
M64(Medium)                                         & 16    & 32    & 48    & 1     & 1     & 8                  & 16  & 64                  & 0.058      & 0.00167   & 0.108  & 0.168  \\
L64(Large)                                          & 16    & 48    & 96    & 1     & 1     & 8                  & 16  & 64                  & 0.166      & 0.00218   & 0.179  & 0.347  \\
G128(Giant)                                         & 16    & 64    & 256   & 1     & 1     & 8                  & 32  & 128                 & 0.809      & 0.00359   & 1.441  & 2.254  \\
E128(Enormous)                                      & 16    & 64    & 256   & 2     & 2     & 8                  & 32  & 128                 & 2.063      & 0.00359   & 1.441  & 3.508  \\
U128(Ultra)                                         & 32    & 128   & 256   & 2     & 2     & 8                  & 32  & 128                 & 2.612      & 0.00423   & 1.784  & 4.400  \\ \hline
\end{tabular}
\label{tab:network-configurations}
\end{table*}

We provide nine model configurations ranging from Atom (A) to Ultra (U) to address diverse application scenarios, as detailed in Table \ref{tab:network-configurations}. The scaling logic for these variants prioritizes increasing capacity in deeper feature layers where spatial dimensions are smaller. Specifically, the 1/2 resolution stage is restricted to a maximum of 32 channels and a single ResNet block to prevent high-resolution layers from becoming speed bottlenecks. Simultaneously, the keypoint detection head maintains a narrow width of 8 channels to maximize throughput on large-scale feature maps. The description head also scales the number of sampling offsets $M$ from 4 up to 32 as the model capacity grows. This architectural approach ensures exceptional efficiency, resulting in a minimal footprint of only 4,252 parameters for the Atom48 variant.

\begin{table}[t]
\centering
\caption{\textbf{Loss weighting configurations.} The table details the coefficients $w_{\mathrm{DS}}$, $w_{\mathrm{OP}}$, and $w_{\mathrm{US}}$ assigned to each model architecture during training.}
\begin{tabular}{lccc}
\hline
\multicolumn{1}{c}{\textbf{Model}} & $w_\mathrm{DS}$ & $w_\mathrm{OP}$ & $w_\mathrm{US}$ \\ \hline
A48                                & 0.05            & 1               & 1               \\
N64                                & 0.1             & 1               & 1               \\
T64                                & 0.5             & 1               & 1               \\
S64                                & 1               & 0               & 1               \\
M64                                & 1               & 0               & 1               \\
L64                                & 1               & 0               & 1               \\
G128                               & 1               & 0               & 1               \\
E128                               & 1               & 0               & 1               \\
U128                               & 1               & 0               & 1               \\ \hline
\end{tabular}
\label{tab:loss-coefficient}
\end{table}

The training objective is defined as a weighted combination of the three loss functions:
\begin{equation}
L_{\mathrm{total}} = w_{\mathrm{DS}} L_{\mathrm{DS}} + w_{\mathrm{OP}} L_{\mathrm{OP}} + w_{\mathrm{US}} L_{\mathrm{US}}.
\end{equation}
The specific weighting coefficients vary according to the model scale as specified in Table~\ref{tab:loss-coefficient}. For the more compact architectures, we utilize the $L_{\mathrm{OP}}$ term to facilitate knowledge distillation. We systematically reduce the weight of the direct metric learning loss $w_{\mathrm{DS}}$ as the model scale decreases to prioritize the distillation signal. This strategy ensures that smaller configurations effectively inherit the representational quality of higher-capacity models.

Our training process utilizes the MegaDepth~\citep{MegaDepth} dataset alongside sampled subsets of ScanNet~\citep{ScanNet} and Revisitop1m~\citep{RevisitOP1m}. We strictly exclude scenes from the MegaDepth-1500 test set to prevent data leakage and ensure fair evaluation. For datasets without ground-truth transformations, we construct training pairs using synthetic homography transformations and apply color augmentation to all 640 $\times$ 640 input images. The network is optimized using the AdamW~\citep{AdamW} optimizer for three epochs with an initial learning rate of 0.002. This learning rate is halved after each epoch to refine model convergence. All model parameters are trained from scratch on four L40S GPUs, with even the largest U128 configuration requiring only one hour to complete.

\begin{table*}
\centering
\caption{\raggedright \textbf{Computation Resources Comparison.} Our methods achieve superior throughput and resource efficiency for edge applications. The best results are marked as \textbf{bold}.}
\begin{tabular}{llcccc}
\hline
                                                    & \multicolumn{1}{c}{\textbf{Method}} & \textbf{Dim} & \textbf{MP}    & \textbf{GFLOPs} & \textbf{FPS}   \\ \hline
\multirow{11}{*}{\rotatebox{90}{\textbf{Standard}}} & SuperPoint~\citep{SuperPoint}       & 256          & 1.301          & 26.11           & 124.2          \\
                                                    & DISK~\citep{DISK}                   & 128          & 1.092          & 98.97           & 26.9           \\
                                                    & ALIKED-T16~\citep{ALIKED}           & 64           & 0.192          & 1.37            & 83.3           \\
                                                    & ALIKED-N16~\citep{ALIKED}           & 128          & 0.677          & 4.05            & 47.7           \\
                                                    & ALIKED-N32~\citep{ALIKED}           & 128          & 0.980          & 4.62            & 46.1           \\
                                                    & AWDesc-T16~\citep{AWDesc}           & 128          & 0.172          & 4.50            & 180.4          \\
                                                    & AWDesc-T32~\citep{AWDesc}           & 128          & 0.390          & 11.81           & 168.2          \\
                                                    & AWDesc-CA~\citep{AWDesc}            & 128          & 10.13          & 27.45           & 84.0           \\
                                                    & DeDoDe-B~\citep{DeDoDev2}           & 256          & 28.08          & 315.58          & 5.27           \\
                                                    & DeDoDe-G~\citep{DeDoDev2}           & 256          & 342.0          & 323.74          & 2.71           \\
                                                    & RDD~\citep{RDD}                     & 256          & 32.90          & 29.25           & 7.31           \\ \hline
\multirow{14}{*}{\rotatebox{90}{\textbf{Fast}}}     & XFeat~\citep{XFeat}                 & 64           & 0.658          & 1.31            & 277.4          \\
                                                    & EdgePoint~\citep{EdgePoint}         & 32           & 0.030          & 0.36            & 838.8          \\
                                                    & EdgePoint2-T48~\citep{EdgePoint2}   & 48           & 0.028          & 0.50            & 503.0          \\
                                                    & EdgePoint2-M64~\citep{EdgePoint2}   & 64           & 0.089          & 1.22            & 437.2          \\
                                                    & EdgePoint2-E64~\citep{EdgePoint2}   & 64           & 0.155          & 1.96            & 375.7          \\
                                                    & Ours-A48                            & 48           & \textbf{0.004} & \textbf{0.08}   & \textbf{881.1} \\
                                                    & Ours-N64                            & 64           & 0.019          & 0.26            & 842.7          \\
                                                    & Ours-T64                            & 64           & 0.043          & 0.29            & 803.5          \\
                                                    & Ours-S64                            & 64           & 0.100          & 0.35            & 746.4          \\
                                                    & Ours-M64                            & 64           & 0.168          & 0.86            & 625.1          \\
                                                    & Ours-L64                            & 64           & 0.347          & 1.03            & 591.4          \\
                                                    & Ours-G128                           & 128          & 2.254          & 2.60            & 475.4          \\
                                                    & Ours-E128                           & 128          & 3.508          & 3.31            & 431.3          \\
                                                    & Ours-U128                           & 128          & 4.400          & 7.12            & 281.4          \\ \hline
\end{tabular}
\label{tab:system-runtime}
\end{table*}

\section{Experiments}

In this section, we first analyze the computational efficiency of our method under the constraints of onboard deployment. We then conduct a comprehensive comparison against state-of-the-art (SOTA) methods across several benchmarks, including homography estimation, relative pose estimation, and visual localization. A series of ablation studies verifies the impact of our architectural design, training strategy, and the efficiency gains from our fused kernel implementation. We also investigate the discriminative power of our descriptors across varying spatial densities to assess robustness under high-density sampling conditions.

We benchmark our approach against a diverse set of SOTA methods, including SuperPoint~\citep{SuperPoint}, DISK~\citep{DISK}, ALIKED~\citep{ALIKED}, AWDesc~\citep{AWDesc}, DeDoDe~\citep{DeDoDe}, and RDD~\citep{RDD}. For the DeDoDe series, we utilize the improved detector from DeDoDe-v2~\citep{DeDoDev2} to generate keypoints. Our evaluation also incorporates efficiency-oriented models such as XFeat~\citep{XFeat}, EdgePoint~\citep{EdgePoint}, and EdgePoint2~\citep{EdgePoint2}.

Unless otherwise specified, we extract 4096 keypoints per image for all evaluated methods to ensure a fair comparison. A Non-Maximum Suppression (NMS) radius of 2 pixels is applied to our method to maintain a balanced spatial distribution. We employ a dual-softmax matcher with a confidence threshold of 0.01 for our models, DeDoDe and RDD, following the setup in~\citep{DeDoDe}. The remaining baseline methods utilize the Mutual-Nearest Neighbor matching strategy as their standard default configuration.

To reach the hardware efficiency targets discussed in Section 3.3, we implement our custom fused kernels in both Triton and CUDA. This specialized implementation enables our method to achieve the reported throughput and latency during evaluation. These optimizations ensure that the computational benefits of our architectural design are fully realized in the practical deployment scenarios analyzed across the following benchmarks.

\subsection{Runtime Efficiency}

\PAR{Setup}
We analyze several efficiency metrics, including descriptor dimensionality (Dim), millions of parameters (MP), and computational complexity (GFLOPs). Furthermore, we evaluate the execution speed on an NVIDIA Jetson Orin-NX, which serves as a standard embedded platform for edge applications. All neural networks are deployed and measured using TensorRT to maximize performance on the integrated GPU. For ALIKED, RDD and our methods, we integrate CUDA acceleration code as TensorRT plugins to ensure equivalent optimization. This assessment focuses on throughput measured in frames per second (FPS) for a standard $480 \times 640$ input resolution with 1024 extracted keypoints, following the protocol in~\citep{ALIKED}.

\PAR{Results}
The results in Table~\ref{tab:system-runtime} demonstrate that our method significantly outperforms all compared baselines in terms of both model compactness and runtime efficiency. Our most lightweight configuration, Ours-A48, requires only 4252 (0.004M) parameters and 0.08 GFLOPs, representing the smallest model size among all tested methods. Despite its minimal resource consumption, it achieves a throughput of over 880 FPS. Even as the model capacity increases, our larger variants maintain competitive performance across the board.

Compared to established lightweight benchmarks like XFeat and EdgePoint2, our models provide a more efficient scaling of descriptor dimensions relative to processing speed. For example, Ours-U128 delivers high-capacity 128-dimensional descriptors while maintaining a frame rate of 281.4 FPS. This throughput is notably higher than many smaller-scale models such as ALIKED and AWDesc. For detection-description decoupled architectures, RDD requires far fewer GFLOPs than DeDoDe yet achieves only slightly higher throughput, due to the inefficiency of its repeated multi-scale dense deformable attention. These results highlight the effectiveness of our architectural design and implementation in maximizing hardware utilization for real-time onboard scenarios.

\subsection{Homography Estimation}

\PAR{Setup}
The HPatches~\citep{HPatches} dataset provides 116 sequences characterized by diverse illumination and viewpoint changes. Following established conventions~\citep{ALIKE,D2-Net}, we exclude eight unreliable scenes and employ mutual nearest neighbor matching. We then utilize MAGSAC++~\citep{MAGSAC} for robust homography estimation between image pairs. Performance is measured using Mean Homography Accuracy (MHA) at error thresholds of 1, 3, and 5 pixels. This metric represents the percentage of correctly projected image corners relative to the ground truth homography.

\begin{table}
\centering
\caption{\raggedright \textbf{Homography estimation on HPatches.} The top three best results in each sector are highlighted in \textcolor{cr}{red}, \textcolor{cg}{green} and \textcolor{cb}{blue}.}
\resizebox{\columnwidth}{!}{%
\begin{tabular}{clccc}
\hline
\multirow{2}{*}{}                                   & \multicolumn{1}{c}{\multirow{2}{*}{\textbf{Method}}} & \multicolumn{3}{c}{\textbf{MHA}}                                                                 \\
                                                    & \multicolumn{1}{c}{}                                 & @1                             & @3                             & @5                             \\ \hline
\multirow{12}{*}{\rotatebox{90}{\textbf{Compact}}}  & ALIKED-T16~\citep{ALIKED}                            & 50.74                          & 83.89                          & 90.56                          \\
                                                    & XFeat~\citep{XFeat}                                  & 47.04                          & 80.19                          & 89.07                          \\
                                                    & EdgePoint~\citep{EdgePoint}                          & 47.96                          & 78.89                          & 87.04                          \\
                                                    & EdgePoint2-T48~\citep{EdgePoint2}                    & 50.19                          & 81.48                          & 88.89                          \\
                                                    & EdgePoint2-M64~\citep{EdgePoint2}                    & 53.70                          & 82.96                          & 89.63                          \\
                                                    & EdgePoint2-E64~\citep{EdgePoint2}                    & 53.33                          & 82.78                          & 90.56                          \\
                                                    & Ours-A48                                             & 49.81                          & 82.22                          & 88.89                          \\
                                                    & Ours-N64                                             & \textcolor{cb}{\textbf{54.63}} & 81.85                          & 88.89                          \\
                                                    & Ours-T64                                             & 52.22                          & 84.26                          & 90.93                          \\
                                                    & Ours-S64                                             & 54.07                          & \textcolor{cr}{\textbf{84.81}} & \textcolor{cg}{\textbf{91.11}} \\
                                                    & Ours-M64                                             & \textcolor{cr}{\textbf{56.11}} & \textcolor{cr}{\textbf{84.81}} & \textcolor{cg}{\textbf{91.11}} \\
                                                    & Ours-L64                                             & \textcolor{cg}{\textbf{55.37}} & \textcolor{cb}{\textbf{84.44}} & \textcolor{cr}{\textbf{92.04}} \\ \hline
\multirow{13}{*}{\rotatebox{90}{\textbf{Standard}}} & SuperPoint~\citep{SuperPoint}                        & 49.81                          & 81.48                          & 88.89                          \\
                                                    & DISK~\citep{DISK}                                    & 51.30                          & 79.44                          & 88.33                          \\
                                                    & ALIKED-N16~\citep{ALIKED}                            & 51.67                          & 83.89                          & 90.37                          \\
                                                    & ALIKED-N32~\citep{ALIKED}                            & 50.37                          & 83.52                          & 90.74                          \\
                                                    & AWDesc-T16~\citep{AWDesc}                            & 53.89                          & 83.33                          & 90.56                          \\
                                                    & AWDesc-T32~\citep{AWDesc}                            & 56.67                          & \textcolor{cg}{\textbf{85.19}} & 90.37                          \\
                                                    & AWDesc-CA~\citep{AWDesc}                             & 54.07                          & \textcolor{cb}{\textbf{85.00}} & \textcolor{cr}{\textbf{92.41}} \\
                                                    & DeDoDe-B~\citep{DeDoDe}                              & 55.56                          & 83.33                          & 89.44                          \\
                                                    & DeDoDe-G~\citep{DeDoDe}                              & 55.93                          & 83.33                          & 90.93                          \\
                                                    & RDD~\citep{RDD}                                      & 46.30                          & 82.96                          & 90.19                          \\
                                                    & Ours-G128                                            & \textcolor{cr}{\textbf{58.15}} & \textcolor{cb}{\textbf{85.00}} & \textcolor{cb}{\textbf{91.11}} \\
                                                    & Ours-E128                                            & \textcolor{cg}{\textbf{57.04}} & 84.63                          & \textcolor{cg}{\textbf{92.04}} \\
                                                    & Ours-U128                                            & \textcolor{cb}{\textbf{56.85}} & \textcolor{cr}{\textbf{86.30}} & 90.74                          \\ \hline
\end{tabular}
}
\label{tab:results-hpatches}
\end{table}

\PAR{Results}
The evaluation on the HPatches dataset in Table \ref{tab:results-hpatches} reveals that our method achieves leading homography accuracy across all error thresholds. Within the compact category, our models consistently outperform established lightweight descriptors such as XFeat, EdgePoint, and variants of EdgePoint2. Notably, even the entry-level Ours-A48 variant exceeds several baselines in precision while utilizing only 0.004 million parameters. This advantage is particularly evident at the strictest 1-pixel threshold, where our configurations maintain significantly higher accuracy under complex conditions.

In the standard category, our high-capacity variants demonstrate a clear performance lead over established benchmarks such as DISK, ALIKED and RDD. Our approach also maintains a competitive edge against significantly heavier frameworks, including DeDoDe and AWDesc. These results demonstrate that our framework achieves superior accuracy while providing a clear efficiency advantage over existing high-performance methods. The consistent ranking of our various configurations across both categories validates the overall reliability of our method for diverse homography estimation tasks.

\subsection{Relative Pose Estimation}

\PAR{Setup}
The MegaDepth-1500~\citep{MegaDepth} and ScanNet-1500~\citep{ScanNet} datasets are employed to assess matching robustness under varying viewpoints and illumination in outdoor and indoor scenes. Additionally, the Image Matching Challenge 2022 (IMC2022)~\citep{IMC2022} serves as a specialized outdoor benchmark for evaluating fundamental matrix estimation. 

Regarding experimental configurations, images are resized to a maximum dimension of 1200 pixels for MegaDepth and $480 \times 640$ for ScanNet. We follow the matching protocols established in~\citep{LoFTR,EfficientLoFTR} and utilize LO-RANSAC~\citep{LO-RANSAC} for essential and fundamental matrix estimation. Optimal matching thresholds for each method are determined according to the strategy in~\citep{XFeat}. Performance is measured using the Area Under the Curve (AUC) for pose accuracy at thresholds of 5$^\circ$, 10$^\circ$, and 20$^\circ$. For IMC2022, images are resized to $800 \times 800$ and we utilize MAGSAC++~\citep{MAGSAC} for robust estimation. The mean Average Accuracy (mAA) is adopted as the primary metric for both private and public subsets.

\begin{table*}
\centering
\caption{\textbf{Relative pose estimation on MegaDepth-1500, ScanNet-1500 and IMC2022.} The top three best results of each sector are marked as \textcolor{cr}{red}, \textcolor{cg}{green} and \textcolor{cb}{blue}.}
\resizebox{\textwidth}{!}{%
\begin{tabular}{clcccccccc}
\hline
\multirow{2}{*}{\textbf{}}                          & \multicolumn{1}{c}{\multirow{2}{*}{\textbf{Method}}} & \multicolumn{3}{c}{\textbf{MegaDepth-1500}}                                                      & \multicolumn{3}{c}{\textbf{ScanNet-1500}}                                                        & \multicolumn{2}{c}{\textbf{IMC2022}}                            \\
                                                    & \multicolumn{1}{c}{}                                 & AUC@5$^\circ$                  & AUC@10$^\circ$                 & AUC@20$^\circ$                 & AUC@5$^\circ$                  & AUC@10$^\circ$                 & AUC@20$^\circ$                 & Private                        & Public                         \\ \hline
\multirow{12}{*}{\rotatebox{90}{\textbf{Compact}}}  & ALIKED-T16~\citep{ALIKED}                            & \textcolor{cr}{\textbf{58.46}} & \textcolor{cr}{\textbf{70.89}} & \textcolor{cg}{\textbf{79.77}} & 14.10                          & 27.18                          & 40.54                          & 0.542                          & 0.559                          \\
                                                    & XFeat~\citep{XFeat}                                  & 43.61                          & 56.78                          & 67.38                          & 12.14                          & 25.08                          & 39.72                          & 0.489                          & 0.505                          \\
                                                    & EdgePoint~\citep{EdgePoint}                          & 42.92                          & 55.72                          & 66.04                          & 11.31                          & 23.46                          & 36.59                          & 0.436                          & 0.460                          \\
                                                    & EdgePoint2-T48~\citep{EdgePoint2}                    & 48.71                          & 60.94                          & 70.67                          & 13.72                          & 27.20                          & 42.01                          & 0.550                          & 0.564                          \\
                                                    & EdgePoint2-M64~\citep{EdgePoint2}                    & 53.20                          & 65.17                          & 74.05                          & 15.28                          & 29.95                          & 44.89                          & \textcolor{cb}{\textbf{0.608}} & \textcolor{cb}{\textbf{0.615}} \\
                                                    & EdgePoint2-E64~\citep{EdgePoint2}                    & 54.32                          & 66.29                          & 75.38                          & 16.20                          & 31.32                          & 46.63                          & \textcolor{cr}{\textbf{0.625}} & \textcolor{cr}{\textbf{0.632}} \\
                                                    & Ours-A48                                             & 44.98                          & 57.13                          & 66.41                          & 12.62                          & 25.35                          & 39.23                          & 0.473                          & 0.492                          \\
                                                    & Ours-N64                                             & 53.91                          & 65.81                          & 75.12                          & 14.88                          & 29.34                          & 44.35                          & 0.583                          & 0.588                          \\
                                                    & Ours-T64                                             & 54.26                          & 66.74                          & 76.16                          & 15.33                          & 31.03                          & 47.37                          & 0.577                          & 0.597                          \\
                                                    & Ours-S64                                             & 55.43                          & 67.74                          & 77.28                          & \textcolor{cb}{\textbf{16.74}} & \textcolor{cb}{\textbf{31.96}} & \textcolor{cb}{\textbf{47.77}} & 0.603                          & 0.605                          \\
                                                    & Ours-M64                                             & \textcolor{cb}{\textbf{57.41}} & \textcolor{cb}{\textbf{69.81}} & \textcolor{cb}{\textbf{78.62}} & \textcolor{cg}{\textbf{17.28}} & \textcolor{cg}{\textbf{33.55}} & \textcolor{cr}{\textbf{49.36}} & 0.602                          & 0.605                          \\
                                                    & Ours-L64                                             & \textcolor{cg}{\textbf{58.29}} & \textcolor{cg}{\textbf{70.83}} & \textcolor{cr}{\textbf{80.02}} & \textcolor{cr}{\textbf{17.86}} & \textcolor{cr}{\textbf{33.59}} & \textcolor{cg}{\textbf{49.33}} & \textcolor{cg}{\textbf{0.619}} & \textcolor{cg}{\textbf{0.617}} \\ \hline
\multirow{13}{*}{\rotatebox{90}{\textbf{Standard}}} & SuperPoint~\citep{SuperPoint}                        & 43.52                          & 56.51                          & 66.78                          & 14.67                          & 28.97                          & 43.76                          & 0.475                          & 0.485                          \\
                                                    & DISK~\citep{DISK}                                    & 54.68                          & 66.63                          & 75.42                          & 13.61                          & 26.31                          & 39.75                          & 0.601                          & 0.589                          \\
                                                    & ALIKED-N16~\citep{ALIKED}                            & 57.62                          & 70.00                          & 79.05                          & 13.16                          & 25.90                          & 39.90                          & 0.630                          & 0.632                          \\
                                                    & ALIKED-N32~\citep{ALIKED}                            & 61.10                          & 73.56                          & 82.16                          & 12.72                          & 24.65                          & 37.05                          & 0.626                          & 0.624                          \\
                                                    & AWDesc-T16~\citep{AWDesc}                            & 48.90                          & 61.27                          & 70.41                          & 14.72                          & 29.41                          & 43.55                          & 0.553                          & 0.574                          \\
                                                    & AWDesc-T32~\citep{AWDesc}                            & 51.26                          & 63.84                          & 72.76                          & 16.79                          & 32.12                          & 47.29                          & 0.583                          & 0.597                          \\
                                                    & AWDesc-CA~\citep{AWDesc}                             & 53.89                          & 66.66                          & 75.87                          & 16.61                          & 32.05                          & 47.00                          & 0.581                          & 0.595                          \\
                                                    & DeDoDe-B~\citep{DeDoDe}                              & 59.56                          & 71.14                          & 79.41                          & 16.75                          & 30.93                          & 44.43                          & 0.623                          & 0.617                          \\
                                                    & DeDoDe-G~\citep{DeDoDe}                              & \textcolor{cb}{\textbf{63.73}} & \textcolor{cb}{\textbf{76.13}} & \textcolor{cb}{\textbf{84.91}} & \textcolor{cr}{\textbf{19.97}} & \textcolor{cr}{\textbf{37.32}} & \textcolor{cg}{\textbf{53.39}} & 0.673                          & 0.667                          \\
                                                    & RDD~\citep{RDD}                                      & \textcolor{cg}{\textbf{64.37}} & \textcolor{cg}{\textbf{77.11}} & \textcolor{cr}{\textbf{86.08}} & \textcolor{cg}{\textbf{19.44}} & \textcolor{cb}{\textbf{36.31}} & \textcolor{cb}{\textbf{52.31}} & \textcolor{cb}{\textbf{0.689}} & \textcolor{cb}{\textbf{0.687}} \\
                                                    & Ours-G128                                            & 62.22                          & 74.61                          & 83.08                          & 18.52                          & 35.04                          & 51.85                          & \textcolor{cb}{\textbf{0.689}} & 0.686                          \\
                                                    & Ours-E128                                            & 62.84                          & 75.16                          & 83.72                          & 18.35                          & 35.38                          & 51.97                          & \textcolor{cg}{\textbf{0.695}} & \textcolor{cg}{\textbf{0.694}} \\
                                                    & Ours-U128                                            & \textcolor{cr}{\textbf{64.75}} & \textcolor{cr}{\textbf{77.15}} & \textcolor{cg}{\textbf{85.77}} & \textcolor{cb}{\textbf{18.94}} & \textcolor{cg}{\textbf{36.46}} & \textcolor{cr}{\textbf{53.66}} & \textcolor{cr}{\textbf{0.717}} & \textcolor{cr}{\textbf{0.716}} \\ \hline
\end{tabular}
}
\label{tab:relative-pose-estimation}
\end{table*}

\begin{figure*}[htbp]
    \centering
    \includegraphics[width=\textwidth]{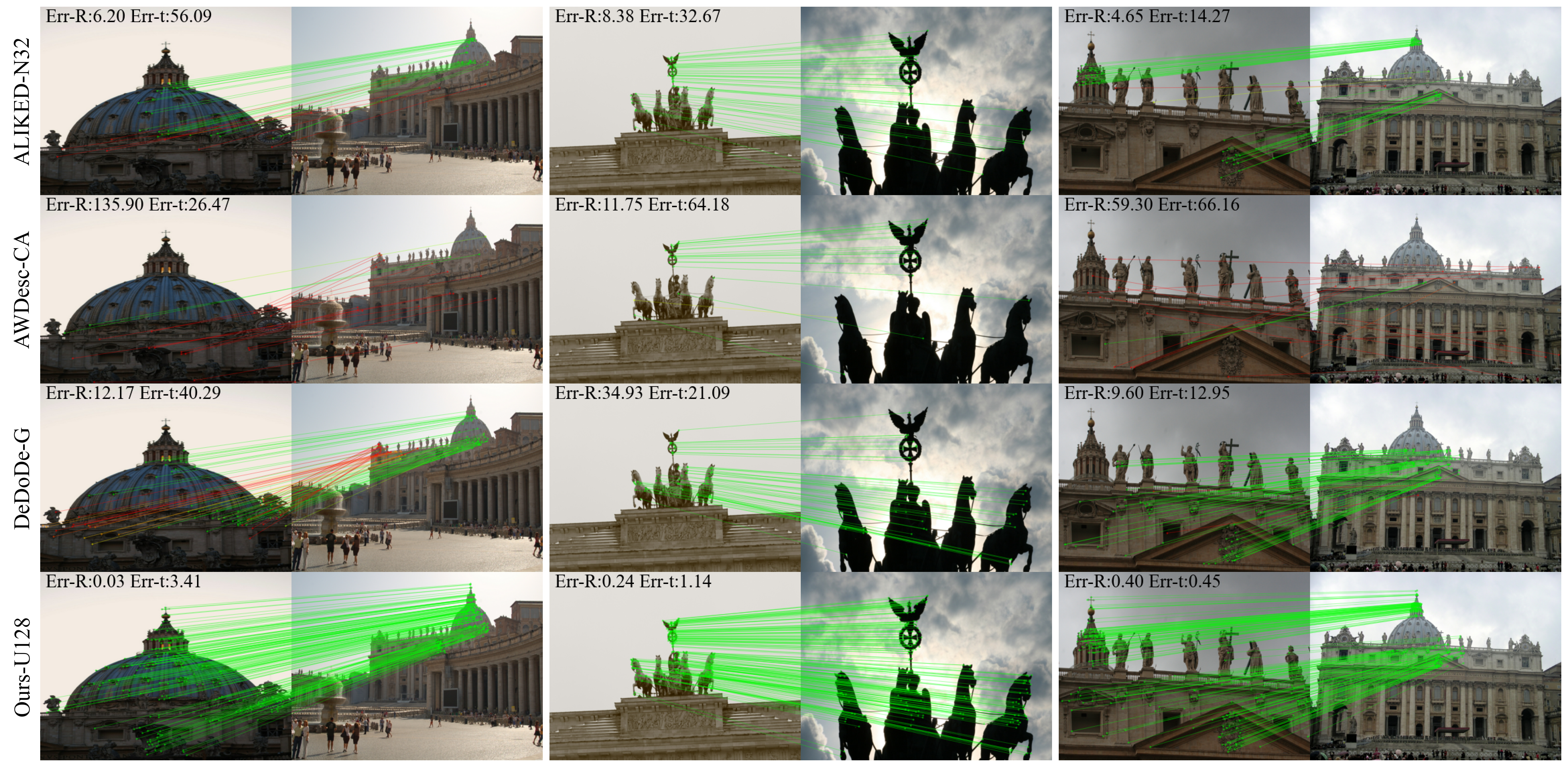}
    \caption{\raggedright \textbf{Qualitative results on MegaDepth-1500.} We compare ALIKED-N32, AWDesc-CA and DeDoDe-G against our U128 model.}
    \label{fig:visualize-megadepth1500}
\end{figure*}

\PAR{Results}
The relative pose estimation results on MegaDepth-1500, ScanNet-1500, and IMC2022 are summarized in Table~\ref{tab:relative-pose-estimation}. For the outdoor MegaDepth-1500 benchmark, our models demonstrate exceptional accuracy across various conditions. In the compact category, our smallest variants remain competitive with state-of-the-art lightweight methods despite their extremely low parameter counts, while our larger compact variants achieve leading accuracy. Among standard descriptors, our high-capacity models achieve the top rankings, surpassing established frameworks such as DeDoDe-B and ALIKED. DeDoDe-G and RDD perform competitively on these benchmarks, though with substantially higher computational cost and descriptor dimensionality. This performance indicates that our features are highly robust to the significant viewpoint and illumination changes typical of outdoor environments, as illustrated in Fig.~\ref{fig:visualize-megadepth1500}.

On the indoor ScanNet-1500 benchmark, our larger compact variants achieve the strongest performance within the compact category. Our U128 model also ranks among the leading standard methods, with only narrow differences separating it from DeDoDe-G and RDD. These results demonstrate that CLIDD remains effective under indoor geometric conditions while retaining a substantial efficiency advantage.

\begin{table*}[!ht]
\centering
\caption{\textbf{Visual localization on Aachen Day-Night v1.1 and InLoc.} The top three best results of each sector are marked as \textcolor{cr}{red}, \textcolor{cg}{green} and \textcolor{cb}{blue}.}
\resizebox{\textwidth}{!}{%
\begin{tabular}{clcccccccccccc}
\hline
\multirow{4}{*}{}                                   & \multicolumn{1}{c}{\multirow{4}{*}{\textbf{Method}}} & \multicolumn{6}{c}{\textbf{Aachen Day-Night v1.1}}                                                                                                                                            & \multicolumn{6}{c}{\textbf{InLoc}}                                                                                                                                                            \\
                                                    & \multicolumn{1}{c}{}                                 & \multicolumn{3}{c}{Day}                                                                       & \multicolumn{3}{c}{Night}                                                                     & \multicolumn{3}{c}{DUC1}                                                                      & \multicolumn{3}{c}{DUC2}                                                                      \\
                                                    & \multicolumn{1}{c}{}                                 & 0.25$m$                       & 0.5$m$                        & 5$m$                          & 0.25$m$                       & 0.5$m$                        & 5$m$                          & 0.25$m$                       & 0.5$m$                        & 5$m$                          & 0.25$m$                       & 0.5$m$                        & 5$m$                          \\
                                                    & \multicolumn{1}{c}{}                                 & 2$^\circ$                     & 5$^\circ$                     & 10$^\circ$                    & 2$^\circ$                     & 5$^\circ$                     & 10$^\circ$                    & 2$^\circ$                     & 5$^\circ$                     & 10$^\circ$                    & 2$^\circ$                     & 5$^\circ$                     & 10$^\circ$                    \\ \hline
\multirow{12}{*}{\rotatebox{90}{\textbf{Compact}}}  & ALIKED-T16~\citep{ALIKED}                            & \textcolor{cg}{\textbf{89.1}} & \textcolor{cg}{\textbf{95.0}} & 98.2                          & \textcolor{cg}{\textbf{77.5}} & \textcolor{cg}{\textbf{91.1}} & \textcolor{cr}{\textbf{99.0}} & 33.8                          & 49.0                          & 59.1                          & 33.6                          & 49.6                          & 57.3                          \\
                                                    & XFeat~\citep{XFeat}                                  & 86.0                          & 93.9                          & 97.6                          & 70.7                          & 85.9                          & 97.9                          & 36.4                          & 54.0                          & 64.1                          & 30.5                          & 44.3                          & 57.3                          \\
                                                    & EdgePoint~\citep{EdgePoint}                          & 86.9                          & 93.2                          & 97.6                          & 66.0                          & 85.9                          & 95.3                          & 28.3                          & 43.4                          & 53.0                          & 22.9                          & 37.4                          & 49.6                          \\
                                                    & EdgePoint2-T48~\citep{EdgePoint2}                    & 86.8                          & 94.1                          & 98.1                          & 74.9                          & 87.4                          & 97.9                          & 34.8                          & 48.5                          & 61.6                          & 37.4                          & 55.7                          & 65.6                          \\
                                                    & EdgePoint2-M64~\citep{EdgePoint2}                    & 87.1                          & 94.7                          & 98.2                          & 75.9                          & 90.1                          & \textcolor{cr}{\textbf{99.0}} & \textcolor{cg}{\textbf{37.9}} & \textcolor{cr}{\textbf{57.1}} & \textcolor{cr}{\textbf{68.7}} & 39.7                          & \textcolor{cg}{\textbf{60.3}} & \textcolor{cb}{\textbf{66.4}} \\
                                                    & EdgePoint2-E64~\citep{EdgePoint2}                    & 86.9                          & \textcolor{cg}{\textbf{95.0}} & \textcolor{cg}{\textbf{98.3}} & \textcolor{cb}{\textbf{77.0}} & \textcolor{cr}{\textbf{91.6}} & \textcolor{cr}{\textbf{99.0}} & 36.4                          & 54.5                          & 66.2                          & \textcolor{cr}{\textbf{41.2}} & \textcolor{cb}{\textbf{58.8}} & \textcolor{cg}{\textbf{67.2}} \\
                                                    & Ours-A48                                             & 86.9                          & 94.5                          & 97.7                          & 72.8                          & 85.3                          & 95.8                          & 28.8                          & 43.9                          & 53.5                          & 28.2                          & 43.5                          & 55.0                          \\
                                                    & Ours-N64                                             & \textcolor{cg}{\textbf{89.1}} & \textcolor{cg}{\textbf{95.0}} & 98.2                          & 74.3                          & 88.5                          & \textcolor{cr}{\textbf{99.0}} & 35.9                          & 50.0                          & 63.1                          & 36.6                          & 52.7                          & 62.6                          \\
                                                    & Ours-T64                                             & 88.7                          & 94.8                          & \textcolor{cg}{\textbf{98.3}} & 72.8                          & 90.1                          & \textcolor{cr}{\textbf{99.0}} & 35.4                          & 55.6                          & 64.6                          & 38.9                          & 52.7                          & 65.6                          \\
                                                    & Ours-S64                                             & 88.8                          & 94.8                          & \textcolor{cr}{\textbf{98.4}} & 75.4                          & 90.1                          & \textcolor{cr}{\textbf{99.0}} & 35.4                          & 52.0                          & 65.2                          & \textcolor{cg}{\textbf{40.5}} & 58.0                          & 65.6                          \\
                                                    & Ours-M64                                             & \textcolor{cr}{\textbf{89.4}} & \textcolor{cr}{\textbf{95.3}} & 98.2                          & 75.9                          & 90.1                          & \textcolor{cr}{\textbf{99.0}} & \textcolor{cb}{\textbf{37.4}} & \textcolor{cb}{\textbf{56.1}} & \textcolor{cb}{\textbf{67.2}} & 39.7                          & 55.0                          & 63.4                          \\
                                                    & Ours-L64                                             & 88.3                          & \textcolor{cg}{\textbf{95.0}} & \textcolor{cg}{\textbf{98.3}} & \textcolor{cr}{\textbf{78.0}} & \textcolor{cb}{\textbf{90.6}} & \textcolor{cr}{\textbf{99.0}} & \textcolor{cr}{\textbf{38.4}} & \textcolor{cg}{\textbf{56.6}} & \textcolor{cg}{\textbf{68.2}} & \textcolor{cg}{\textbf{40.5}} & \textcolor{cr}{\textbf{61.1}} & \textcolor{cr}{\textbf{67.9}} \\ \hline
\multirow{13}{*}{\rotatebox{90}{\textbf{Standard}}} & SuperPoint~\citep{SuperPoint}                        & 88.3                          & 94.4                          & 98.1                          & 69.1                          & 85.9                          & 95.8                          & 33.3                          & 50.0                          & 59.6                          & 32.8                          & 51.9                          & 63.4                          \\
                                                    & DISK~\citep{DISK}                                    & 87.3                          & 95.5                          & 98.5                          & \textcolor{cg}{\textbf{78.0}} & 89.0                          & 99.0                          & 35.9                          & 53.0                          & 66.2                          & 24.4                          & 40.5                          & 57.3                          \\
                                                    & ALIKED-N16~\citep{ALIKED}                            & 88.8                          & \textcolor{cr}{\textbf{96.1}} & \textcolor{cg}{\textbf{98.8}} & 74.3                          & 89.5                          & 99.0                          & 32.8                          & 54.0                          & 65.7                          & 35.1                          & 50.4                          & 58.0                          \\
                                                    & ALIKED-N32~\citep{ALIKED}                            & 86.4                          & 95.3                          & 98.3                          & 74.9                          & 90.1                          & \textcolor{cr}{\textbf{99.5}} & 36.9                          & 51.0                          & 66.7                          & 35.9                          & 49.6                          & 64.1                          \\
                                                    & AWDesc-T16~\citep{AWDesc}                            & \textcolor{cg}{\textbf{89.2}} & 95.1                          & 98.4                          & 75.4                          & 88.0                          & 96.9                          & 38.9                          & 56.1                          & 66.7                          & 35.1                          & 51.9                          & 64.9                          \\
                                                    & AWDesc-T32~\citep{AWDesc}                            & 88.2                          & \textcolor{cg}{\textbf{95.9}} & 98.4                          & \textcolor{cb}{\textbf{77.0}} & 90.1                          & 98.4                          & 38.4                          & 57.6                          & 68.7                          & 34.4                          & 55.7                          & 65.6                          \\
                                                    & AWDesc-CA~\citep{AWDesc}                             & 88.5                          & 95.4                          & 98.4                          & 74.9                          & \textcolor{cr}{\textbf{91.1}} & 99.0                          & 39.9                          & 60.1                          & 71.2                          & 42.7                          & 58.0                          & 68.7                          \\
                                                    & DeDoDe-B~\citep{DeDoDe}                              & 86.5                          & 95.0                          & 98.2                          & 74.9                          & 89.0                          & \textcolor{cr}{\textbf{99.5}} & 32.8                          & 55.1                          & 69.2                          & 33.6                          & 47.3                          & 57.3                          \\
                                                    & DeDoDe-G~\citep{DeDoDe}                              & 88.0                          & 95.4                          & 98.5                          & 74.3                          & 90.6                          & \textcolor{cr}{\textbf{99.5}} & 40.9                          & \textcolor{cg}{\textbf{63.6}} & \textcolor{cg}{\textbf{74.7}} & 42.0                          & 58.0                          & 65.6                          \\
                                                    & RDD~\citep{RDD}                                      & 88.8                          & 95.6                          & \textcolor{cr}{\textbf{98.9}} & 73.8                          & \textcolor{cr}{\textbf{91.1}} & \textcolor{cr}{\textbf{99.5}} & \textcolor{cr}{\textbf{45.5}} & \textcolor{cr}{\textbf{65.2}} & \textcolor{cr}{\textbf{77.8}} & \textcolor{cr}{\textbf{49.6}} & \textcolor{cr}{\textbf{65.6}} & \textcolor{cr}{\textbf{72.0}} \\
                                                    & Ours-G128                                            & 88.8                          & 95.6                          & 98.7                          & 75.4                          & 90.1                          & 99.0                          & \textcolor{cb}{\textbf{41.9}} & 61.1                          & 71.2                          & \textcolor{cg}{\textbf{47.3}} & \textcolor{cg}{\textbf{64.1}} & \textcolor{cb}{\textbf{69.5}} \\
                                                    & Ours-E128                                            & \textcolor{cb}{\textbf{89.0}} & 95.6                          & 98.7                          & \textcolor{cr}{\textbf{78.5}} & 90.6                          & 99.0                          & 41.4                          & 59.6                          & 70.2                          & 45.0                          & \textcolor{cg}{\textbf{64.1}} & \textcolor{cg}{\textbf{70.2}} \\
                                                    & Ours-U128                                            & \textcolor{cr}{\textbf{89.4}} & \textcolor{cb}{\textbf{95.8}} & \textcolor{cg}{\textbf{98.8}} & 76.4                          & \textcolor{cr}{\textbf{91.1}} & 99.0                          & \textcolor{cg}{\textbf{43.4}} & \textcolor{cb}{\textbf{62.1}} & \textcolor{cb}{\textbf{71.7}} & \textcolor{cb}{\textbf{45.8}} & 61.1                          & 67.9                          \\ \hline
\end{tabular}%
}
\label{tab:hloc}
\end{table*}

The evaluation on IMC2022 further confirms the superiority of our framework for fundamental matrix estimation. Our standard models achieve the highest mean Average Accuracy on both private and public subsets, notably outperforming previous state-of-the-art results from DeDoDe-G and RDD. Similarly, our compact variants maintain a high ranking, with many configurations rivaling the performance of standard-sized descriptors. This outcome validates the high discriminative power of our features even under the rigorous requirements of large-scale competition benchmarks.

Overall, our framework achieves state-of-the-art performance on the outdoor evaluations and ranks among the strongest methods indoors, while providing a substantial efficiency advantage. Using a standard CNN architecture, our U128 model delivers performance comparable to or better than the DINOv2-based DeDoDe-G, with a compact 128-dimensional descriptor and approximately 1.3\% of its parameters. Simultaneously, our ultra-lightweight A48 variant achieves matching performance comparable to SuperPoint while utilizing only approximately 0.3\% of its parameter count. These results confirm that our architecture delivers leading matching accuracy with the exceptional efficiency required for real-time deployment.

\subsection{Visual Localization}

\PAR{Setup} 
The visual localization performance is evaluated using the Aachen Day-Night v1.1~\citep{HLocDataset} and InLoc~\citep{InLoc} datasets to assess robustness against extreme day-night illumination changes and complex indoor environments. We integrate our descriptors into the Hierarchical Localization framework~\citep{HLoc} following the standard pipeline for feature extraction and matching. In accordance with recommended protocols, images are resized to a maximum dimension of 1024 pixels for Aachen and 1600 pixels for InLoc. Accuracy is quantified by the percentage of successfully localized images within translation and rotation thresholds of (0.25m, 2$^\circ$), (0.5m, 5$^\circ$), and (5m, 10$^\circ$).

\PAR{Results}
The visual localization results are presented in Table~\ref{tab:hloc}. On the Aachen Day-Night benchmark, our framework consistently achieves top-tier accuracy across both day and night conditions. Our compact models provide robustness that rivals many standard-sized methods. Meanwhile, our standard variants remain highly competitive against established pipelines such as ALIKED and DISK. These results indicate that our descriptors are exceptionally stable under the extreme illumination transitions characteristic of day-night cycles.

For the indoor InLoc dataset, our approach demonstrates a clear performance advantage in environments with significant viewpoint changes. In the standard sector, our models occupy leading positions across nearly all thresholds for both DUC1 and DUC2 subsets. This performance notably exceeds other high-capacity methods like AWDesc and DeDoDe-B, the latter of which shows more limited generalization to these indoor scenes. RDD leads on InLoc but operates with larger descriptors and markedly lower throughput than our framework. The superior results on InLoc underscore the high discriminative power and geometric consistency of our learned features.

Overall, these experiments demonstrate that our framework delivers top-tier localization accuracy across diverse and challenging environments while maintaining the real-time efficiency required for practical applications. This consistent performance across outdoor illumination changes and indoor geometric variations validates the effectiveness of our approach. Our framework remains ideal for practical applications requiring both high reliability and real-time processing speeds.

\subsection{Ablation Study}

\subsubsection{Design of the CLIDD}

\begin{table*}[ht!]
\centering
\caption{\raggedright \textbf{Ablation study of specific structural configurations.} These experiments compare single-layer versus cross-layer aggregation for offset prediction and shared versus layer-independent strategies for feature sampling.}
\begin{tabular}{ccccccccc}
\hline
\multirow{2}{*}{\textbf{Model}} & \multirow{2}{*}{\textbf{Predictor}} & \multirow{2}{*}{\textbf{Sampler}} & \multicolumn{3}{c}{\textbf{MegaDepth-1500}}      & \multicolumn{3}{c}{\textbf{ScanNet-1500}}        \\
                                &                                     &                                   & AUC@5$^\circ$  & AUC@10$^\circ$ & AUC@20$^\circ$ & AUC@5$^\circ$  & AUC@10$^\circ$ & AUC@20$^\circ$ \\ \hline
\multirow{3}{*}{A48}            & Single                              & Independent                       & 43.99          & 56.20          & 65.89          & 12.41          & 25.19          & 38.92          \\
                                & Cross                               & Independent                       & \textbf{44.98} & \textbf{57.13} & \textbf{66.41} & \textbf{12.62} & \textbf{25.35} & \textbf{39.23} \\
                                & Cross                               & Shared                            & 43.20          & 55.22          & 64.72          & 12.15          & 24.41          & 37.82          \\ \hdashline
\multirow{3}{*}{U128}           & Single                              & Independent                       & 63.17          & 75.51          & 84.14          & 18.69          & 35.91          & 52.74          \\
                                & Cross                               & Independent                       & \textbf{64.75} & \textbf{77.15} & \textbf{85.77} & \textbf{18.94} & \textbf{36.46} & \textbf{53.66} \\
                                & Cross                               & Shared                            & 63.43          & 75.80          & 84.38          & 18.14          & 35.04          & 52.02          \\ \hline
\end{tabular}
\label{tab:description-head-sparse}
\end{table*}

We evaluate the internal components of CLIDD by analyzing the design of the Predictor and Sampler. As shown in Table~\ref{tab:description-head-sparse}, our sparse extraction strategy is compared against several internal baselines using the MegaDepth-1500 and ScanNet-1500 benchmarks. The results demonstrate that the Cross-Layer Predictor outperforms configurations utilizing only a single feature resolution. Similarly, the Layer-Independent Sampler yields higher precision than coupled sampling alternatives. Integrating both components maximizes geometric robustness and discriminative power. These gains remain consistent across both the lightweight A48 and high-capacity U128 variants, confirming the architectural scalability of our approach.

\subsubsection{Comparison with Dense Description Head}

\begin{table*}
\centering
\caption{\raggedright \textbf{Comparison with dense description architectures.} The experimental results evaluate our sparse extraction strategy against several dense fusion description method.}
\resizebox{\textwidth}{!}{
\begin{tabular}{clccccccc}
\hline
\multirow{2}{*}{\textbf{Model}} & \multicolumn{1}{c}{\multirow{2}{*}{\textbf{\begin{tabular}[c]{@{}c@{}}Description\\ Architecture\end{tabular}}}} & \multirow{2}{*}{\textbf{FPS}} & \multicolumn{3}{c}{\textbf{MegaDepth-1500}}      & \multicolumn{3}{c}{\textbf{ScanNet-1500}}        \\
                                & \multicolumn{1}{c}{}                                                                                             &                               & AUC@5$^\circ$  & AUC@10$^\circ$ & AUC@20$^\circ$ & AUC@5$^\circ$  & AUC@10$^\circ$ & AUC@20$^\circ$ \\ \hline
\multirow{4}{*}{A48}            & Sub-Scale Vanilla~\citep{EdgePoint2}                                                                             & 845.3                         & 25.28          & 36.39          & 47.46          & 9.18           & 18.89          & 30.77          \\
                                & Full-Scale Vanilla~\citep{DeDoDe}                                                                                & 420.2                         & 13.47          & 21.15          & 30.53          & 5.12           & 12.06          & 21.90          \\
                                & Full-Scale SDDH~\citep{ALIKED}                                                                                   & 162.4                         & 42.56          & 54.27          & 63.93          & 11.42          & 23.39          & 36.66          \\
                                & Ours                                                                                                             & 881.1                         & \textbf{44.98} & \textbf{57.13} & \textbf{66.41} & \textbf{12.62} & \textbf{25.35} & \textbf{39.23} \\ \hdashline
\multirow{4}{*}{U128}           & Sub-Scale Vanilla~\citep{EdgePoint2}                                                                             & 170.2                         & 60.82          & 73.07          & 81.94          & 18.28          & 35.00          & 51.09          \\
                                & Full-Scale Vanilla~\citep{DeDoDe}                                                                                & 164.6                         & 56.16          & 67.89          & 76.65          & 17.33          & 33.91          & 49.87          \\
                                & Full-Scale SDDH~\citep{ALIKED}                                                                                   & 64.6                          & 63.28          & 75.63          & 84.51          & 17.93          & 33.91          & 51.45          \\
                                & Ours                                                                                                             & 281.4                         & \textbf{64.75} & \textbf{77.15} & \textbf{85.77} & \textbf{18.94} & \textbf{36.46} & \textbf{53.66} \\ \hline
\end{tabular}
}
\label{tab:description-head-dense}
\end{table*}

We evaluate CLIDD against description methods requiring dense feature fusion. As shown in Table~\ref{tab:description-head-dense}, both sub-scale and full-scale vanilla strategies exhibit limited effectiveness when applied to the ultra-lightweight A48 backbone. While scaling up to the high-capacity U128 model improves accuracy, the results remain inferior to our CLIDD-based approach. We also observe that the full-scale strategy imposes significant memory pressure during training, requiring more than double the GPU memory per card compared to our method. To accommodate these intensive resource demands, we reduce the training batch size to one-fourth of the original configuration and employ gradient accumulation. Despite these adjustments, the full-scale vanilla method still underperforms relative to the sub-scale version. This performance degradation likely stems from the shallow design of our backbone compared to the heavy architectures utilized in~\citep{DeDoDe}. Furthermore, while the deformable-based SDDH outperforms vanilla baselines, it fails to match our method in precision and inference speed. Notably, even with kernel fusion acceleration, the full-scale SDDH remains substantially less efficient than our framework. These results confirm the superior effectiveness and efficiency of CLIDD for high-quality local feature representation.

To further validate the effectiveness of CLIDD in utilizing multi-scale features produced by an existing backbone, we integrate CLIDD into three pretrained ALIKED variants: T16, N16, and N32. SDDH and CLIDD both employ deformable sampling, but SDDH first fuses the four ALIKED feature scales into a unified dense map, whereas CLIDD samples them directly and independently. During training, the pretrained ALIKED backbone and detection branch remain frozen, and only the newly introduced CLIDD parameters are optimized using the DualSoftmax loss, without descriptor distillation. As shown in Table~\ref{tab:description-head-with-aliked}, CLIDD consistently outperforms SDDH across all three ALIKED variants and all evaluation thresholds. These experiments on fixed pretrained ALIKED features provide complementary evidence that CLIDD's direct and independent multi-scale sampling can improve descriptor performance over the dense-fusion description head used by SDDH.

\begin{table*}
\centering
\caption{\raggedright \textbf{Comparison of description architectures on frozen ALIKED models.} The ALIKED backbone and detection branch are kept frozen, while the original SDDH is replaced with CLIDD and only the CLIDD parameters are optimized.}
\begin{tabular}{clcccccc}
\hline
\multirow{2}{*}{\textbf{\begin{tabular}[c]{@{}c@{}}ALIKED\\ Model\end{tabular}}} & \multicolumn{1}{c}{\multirow{2}{*}{\textbf{\begin{tabular}[c]{@{}c@{}}Description\\ Architecture\end{tabular}}}} & \multicolumn{3}{c}{\textbf{MegaDepth-1500}}      & \multicolumn{3}{c}{\textbf{ScanNet-1500}}        \\
                                                                                 & \multicolumn{1}{c}{}                                                                                             & AUC@5$^\circ$  & AUC@10$^\circ$ & AUC@20$^\circ$ & AUC@5$^\circ$  & AUC@10$^\circ$ & AUC@20$^\circ$ \\ \hline
\multirow{2}{*}{T16}                                                             & SDDH~\citep{ALIKED}                                                                                              & 58.46          & 70.89          & 79.77          & 14.10          & 27.18          & 40.54          \\
                                                                                 & CLIDD                                                                                                            & \textbf{58.73} & \textbf{71.51} & \textbf{80.48} & \textbf{15.07} & \textbf{29.49} & \textbf{44.17} \\ \hdashline
\multirow{2}{*}{N16}                                                             & SDDH~\citep{ALIKED}                                                                                              & 57.62          & 70.00          & 79.05          & 13.16          & 25.90          & 39.90          \\
                                                                                 & CLIDD                                                                                                            & \textbf{59.24} & \textbf{71.46} & \textbf{80.31} & \textbf{15.05} & \textbf{28.89} & \textbf{43.36} \\ \hdashline
\multirow{2}{*}{N32}                                                             & SDDH~\citep{ALIKED}                                                                                              & 61.10          & 73.56          & 82.16          & 12.72          & 24.65          & 37.05          \\
                                                                                 & CLIDD                                                                                                            & \textbf{61.75} & \textbf{74.40} & \textbf{83.30} & \textbf{13.60} & \textbf{26.27} & \textbf{39.59} \\ \hline
\end{tabular}
\label{tab:description-head-with-aliked}
\end{table*}

\subsubsection{Descriptor Distillation}

\begin{table*}
\centering
\caption{\raggedright \textbf{Ablation of descriptor loss weighting.} Performance comparison across different weight balances for the DualSoftmax loss and the Orthogonal-Procrustes loss.}
\begin{tabular}{ccccccccc}
\hline
\multirow{2}{*}{\textbf{Model}} & \multirow{2}{*}{$w_{\mathrm{DS}}$} & \multirow{2}{*}{$w_{\mathrm{OP}}$} & \multicolumn{3}{c}{\textbf{MegaDepth-1500}}      & \multicolumn{3}{c}{\textbf{ScanNet-1500}}        \\
                                &                                    &                                    & AUC@5$^\circ$  & AUC@10$^\circ$ & AUC@20$^\circ$ & AUC@5$^\circ$  & AUC@10$^\circ$ & AUC@20$^\circ$ \\ \hline
\multirow{4}{*}{A48}            & 0                                  & 1                                  & 43.98          & 55.93          & 65.17          & 11.86          & 23.91          & 37.09          \\
                                & 0.05                               & 1                                  & \textbf{44.98} & \textbf{57.13} & \textbf{66.41} & \textbf{12.62} & \textbf{25.35} & \textbf{39.23} \\
                                & 0.1                                & 1                                  & 42.30          & 54.93          & 65.25          & 11.16          & 23.37          & 37.56          \\
                                & 1                                  & 0                                  & 33.87          & 46.65          & 57.99          & 9.86           & 21.12          & 34.05          \\ \hdashline
\multirow{5}{*}{N64}            & 0                                  & 1                                  & 52.71          & 64.21          & 72.80          & 13.97          & 27.83          & 42.01          \\
                                & 0.05                               & 1                                  & 52.51          & 64.40          & 73.11          & 14.13          & 27.82          & 42.21          \\
                                & 0.1                                & 1                                  & \textbf{53.91} & \textbf{65.81} & \textbf{75.12} & \textbf{14.88} & \textbf{29.34} & \textbf{44.35} \\
                                & 0.5                                & 1                                  & 53.62          & 65.53          & 74.31          & 14.50          & 29.14          & 44.24          \\
                                & 1                                  & 0                                  & 50.49          & 62.87          & 71.98          & 14.93          & 29.25          & 44.27          \\ \hdashline
\multirow{5}{*}{T64}            & 0                                  & 1                                  & 53.55          & 65.73          & 74.54          & 14.86          & 28.64          & 43.00          \\
                                & 0.1                                & 1                                  & \textbf{55.54} & \textbf{67.47} & 76.27          & 15.24          & 29.36          & 44.06          \\
                                & 0.5                                & 1                                  & 54.26          & 66.74          & 76.16          & 15.33          & \textbf{31.03} & \textbf{47.37} \\
                                & 1                                  & 1                                  & 54.53          & 67.15          & \textbf{76.32} & \textbf{15.63} & 30.65          & 46.18          \\
                                & 1                                  & 0                                  & 53.94          & 66.44          & 75.57          & 15.49          & 30.72          & 46.27          \\ \hline
\end{tabular}
\label{tab:description-loss}
\end{table*}

\begin{table*}
\centering
\caption{\raggedright \textbf{Ablation of teacher model for descriptor distillation.} Comparison of descriptor performance with ALIKED-N32 \textit{vs.} our U128 model as the distillation teacher.} 
\begin{tabular}{cccccccc}
\hline
\multirow{2}{*}{\textbf{Model}} & \multirow{2}{*}{\textbf{\begin{tabular}[c]{@{}c@{}}Teacher\\ Description\end{tabular}}} & \multicolumn{3}{c}{\textbf{MegaDepth-1500}}      & \multicolumn{3}{c}{\textbf{ScanNet-1500}}        \\
                                &                                                                                         & AUC@5$^\circ$  & AUC@10$^\circ$ & AUC@20$^\circ$ & AUC@5$^\circ$  & AUC@10$^\circ$ & AUC@20$^\circ$ \\ \hline
\multirow{2}{*}{A48}            & ALIKED-N32                                                                              & 44.98          & 57.13          & 66.41          & \textbf{12.62} & \textbf{25.35} & \textbf{39.23} \\
                                & Ours-U128                                                                               & \textbf{46.43} & \textbf{58.53} & \textbf{67.87} & 12.56          & 25.03          & 38.93          \\ \hdashline
\multirow{2}{*}{N64}            & ALIKED-N32                                                                              & 53.91          & \textbf{65.81} & \textbf{75.12} & 14.88          & \textbf{29.34} & \textbf{44.35} \\
                                & Ours-U128                                                                               & \textbf{54.06} & 65.79          & 74.55          & \textbf{14.97} & 29.25          & 44.04          \\ \hdashline
\multirow{2}{*}{T64}            & ALIKED-N32                                                                              & 54.26          & 66.74          & \textbf{76.16} & 15.33          & 31.03          & \textbf{47.37} \\
                                & Ours-U128                                                                               & \textbf{55.14} & \textbf{67.06} & 76.11          & \textbf{16.07} & \textbf{31.45} & 47.24          \\ \hline
\end{tabular}
\label{tab:description-teacher}
\end{table*}

We examine the influence of descriptor loss components by testing various weight combinations of the DualSoftmax loss $L_{\mathrm{DS}}$ for direct supervision and the Orthogonal-Procrustes loss $L_{\mathrm{OP}}$ for description distillation. This experiment targets lightweight architectures to determine how different supervision strategies affect feature discriminability under capacity constraints. By evaluating configurations ranging from pure distillation to pure direct training, we identify the optimal weighting required to achieve high accuracy in small-scale models. A further experiment investigates whether our own U128 model can substitute for ALIKED-N32 as the distillation teacher, testing whether descriptor-level supervision from a larger variant matches that of an external teacher.

We first examine the loss weight ablation in Table~\ref{tab:description-loss}. The distillation loss proves vital for the most lightweight architectures. For the minimalist A48 variant, the highest accuracy is achieved when the direct supervision weight $w_{\mathrm{DS}}$ is reduced to 0.05. Relying solely on the direct metric loss leads to a significant performance drop for this architecture. As model capacity grows, the optimal weight for direct supervision gradually increases. For instance, the N64 and T64 models reach peak performance with $w_{\mathrm{DS}}$ values of 0.1 and 0.5, respectively. However, the performance improvements from distillation for the T64 model are relatively limited compared to those observed for the A48 and N64 variants. This trend suggests that higher-capacity models can effectively leverage ground-truth signals with less reliance on external teacher guidance. Nevertheless, a balanced combination of both losses consistently outperforms single-loss strategies across all tested scales. These findings confirm that mixed supervision is critical for maximizing the potential of lightweight descriptor models.

Table~\ref{tab:description-teacher} compares ALIKED-N32 and our U128 model as distillation teachers. Substituting U128 for ALIKED-N32 produces comparable performance across all three student variants. On MegaDepth-1500, U128-guided models yield slightly higher AUC scores. On ScanNet-1500, the two teachers produce comparable scores. These results show that descriptor distillation benefits the compact variants but is not tied to ALIKED, since a larger CLIDD model provides comparable supervision.

\subsubsection{Keypoint Position Prior}

During training, the pretrained teacher model serves two roles: it supplies keypoint positions as detection supervision targets and also as descriptor sampling locations. To test whether CLIDD's descriptor and detector generalize across different teacher signals, we decouple these two roles and independently substitute each one with an alternative source. We train the U128 model under four configurations, independently substituting ALIKED-N32 with SuperPoint for each role.

\begin{table*}
\centering
\caption{\raggedright \textbf{Ablation on keypoint position prior.} Performance comparison when substituting ALIKED-N32 with SuperPoint for detection supervision and descriptor sampling locations independently.}
\begin{tabular}{cccccccc}
\hline
\multicolumn{2}{c}{\textbf{Keypoint Prior}} & \multicolumn{3}{c}{\textbf{MegaDepth-1500}}      & \multicolumn{3}{c}{\textbf{ScanNet-1500}}        \\
Detection             & Description           & AUC@5$^\circ$  & AUC@10$^\circ$ & AUC@20$^\circ$ & AUC@5$^\circ$  & AUC@10$^\circ$ & AUC@20$^\circ$ \\ \hline
ALIKED-N32            & ALIKED-N32            & 64.75          & \textbf{77.15} & \textbf{85.77} & 18.94          & 36.46          & 53.66          \\
ALIKED-N32            & SuperPoint            & \textbf{64.85} & 77.03          & 85.47          & \textbf{19.00} & 35.94          & 53.19          \\
SuperPoint            & ALIKED-N32            & 60.13          & 72.90          & 82.15          & 18.89          & \textbf{36.62} & \textbf{54.31} \\
SuperPoint            & SuperPoint            & 59.57          & 72.67          & 82.14          & 18.68          & 35.43          & 52.20          \\ \hline
\end{tabular}
\label{tab:ablation-keypoint-prior}
\end{table*}

\begin{figure*}[!htbp]
    \centering
    \includegraphics[width=0.75\textwidth]{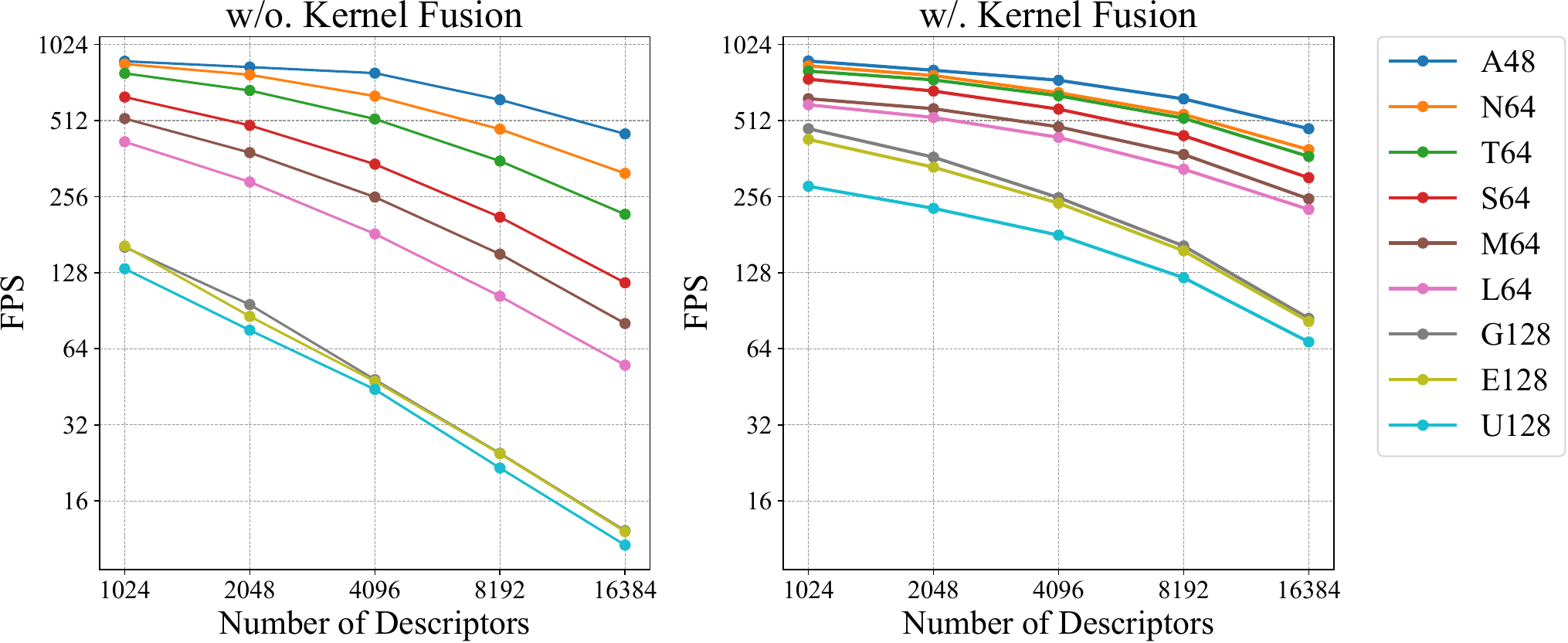}
    \caption{\raggedright \textbf{Throughput comparison across various model scales and keypoint counts.} The results visualize the inference efficiency with and without kernel fusion. Both the number of sampling points and the FPS results are presented on a $\log_2$ scale.}
    \label{fig:kernel_fusion}
\end{figure*}

Table~\ref{tab:ablation-keypoint-prior} reports the results. When SuperPoint provides the descriptor sampling positions while ALIKED-N32 remains the detection teacher, performance remains nearly identical to using ALIKED-N32 for both roles. This indicates that descriptor learning does not depend specifically on ALIKED; a method that yields usable keypoint locations is generally adequate. When SuperPoint replaces ALIKED as the detection teacher, a clear performance drop occurs on MegaDepth-1500, whereas ScanNet-1500 shows only marginal change. This pattern matches the outdoor-indoor performance profile of the two teachers. ALIKED holds a stronger advantage outdoors, while the two methods perform comparably indoors. Detection quality therefore reflects the capability of its supervising teacher, which is consistent with the expected behavior of distillation-based training. These results confirm that CLIDD's descriptor learning is independent of the specific teacher.

\subsubsection{Efficiency of Kernel Fusion}

\begin{table*}[!htbp]
\centering
\caption{\raggedright \textbf{Evaluation of descriptor discriminability under varying spatial densities.} Comparison of matching performance across different keypoint limits and NMS configurations for multiple model scales.}
\resizebox{0.9\textwidth}{!}{
\begin{tabular}{ccccccccccc}
\hline
\multirow{2}{*}{\textbf{Method}} & \multicolumn{2}{c}{\textbf{Configuration}} & \multicolumn{3}{c}{\textbf{MegaDepth-1500}}      & \multicolumn{3}{c}{\textbf{ScanNet-1500}}        & \multicolumn{2}{c}{\textbf{IMC2022}} \\
                                 & Max $N$        & NMS                & AUC@5$^\circ$  & AUC@10$^\circ$ & AUC@20$^\circ$ & AUC@5$^\circ$  & AUC@10$^\circ$ & AUC@20$^\circ$ & Private           & Public           \\ \hline
\multirow{4}{*}{A48}             & 1024             & $\checkmark$       & 39.08          & 51.52          & 61.76          & 12.39          & 24.47          & 37.90          & 0.402             & 0.417            \\
                                 & 4096             & $\checkmark$       & 44.98          & 57.13          & 66.41          & 12.62          & 25.35          & 39.23          & 0.473             & 0.492            \\
                                 & 4096             & $\times$           & 48.84          & 61.06          & 70.32          & 14.45          & 28.03          & 42.31          & 0.493             & 0.511            \\
                                 & 10000            & $\times$           & \textbf{51.96} & \textbf{63.65} & \textbf{72.26} & \textbf{15.37} & \textbf{29.86} & \textbf{44.31} & \textbf{0.530}    & \textbf{0.548}   \\ \hdashline
\multirow{4}{*}{N64}             & 1024             & $\checkmark$       & 48.26          & 60.69          & 70.38          & 14.50          & 28.25          & 42.51          & 0.515             & 0.520            \\
                                 & 4096             & $\checkmark$       & 53.91          & 65.81          & 75.12          & 14.88          & 29.34          & 44.35          & 0.583             & 0.588            \\
                                 & 4096             & $\times$           & 56.62          & 68.69          & 77.24          & 16.33          & 31.63          & 46.87          & 0.601             & 0.612            \\
                                 & 10000            & $\times$           & \textbf{58.43} & \textbf{70.46} & \textbf{79.08} & \textbf{17.30} & \textbf{32.99} & \textbf{48.09} & \textbf{0.636}    & \textbf{0.644}   \\ \hdashline
\multirow{4}{*}{T64}             & 1024             & $\checkmark$       & 49.15          & 62.34          & 72.48          & 15.19          & 30.41          & 45.56          & 0.509             & 0.518            \\
                                 & 4096             & $\checkmark$       & 54.26          & 66.74          & 76.16          & 15.33          & 31.03          & 47.37          & 0.577             & 0.597            \\
                                 & 4096             & $\times$           & 57.01          & 69.43          & 78.56          & 17.05          & 33.07          & 48.92          & 0.604             & 0.613            \\
                                 & 10000            & $\times$           & \textbf{59.17} & \textbf{71.53} & \textbf{80.23} & \textbf{17.70} & \textbf{33.83} & \textbf{49.79} & \textbf{0.624}    & \textbf{0.641}   \\ \hdashline
\multirow{4}{*}{S64}             & 1024             & $\checkmark$       & 49.80          & 63.42          & 73.67          & 16.03          & 30.90          & 46.34          & 0.513             & 0.521            \\
                                 & 4096             & $\checkmark$       & 55.43          & 67.74          & 77.28          & 16.74          & 31.96          & 47.77          & 0.603             & 0.605            \\
                                 & 4096             & $\times$           & 57.16          & 70.34          & 80.04          & 17.90          & 33.72          & 48.75          & 0.616             & 0.620            \\
                                 & 10000            & $\times$           & \textbf{59.73} & \textbf{72.48} & \textbf{81.63} & \textbf{18.66} & \textbf{34.84} & \textbf{50.00} & \textbf{0.642}    & \textbf{0.653}   \\ \hdashline
\multirow{4}{*}{M64}             & 1024             & $\checkmark$       & 49.36          & 62.31          & 72.41          & 16.53          & 32.27          & 48.11          & 0.533             & 0.528            \\
                                 & 4096             & $\checkmark$       & 57.41          & 69.81          & 78.62          & 17.28          & 33.55          & 49.36          & 0.602             & 0.605            \\
                                 & 4096             & $\times$           & 58.99          & 71.70          & 80.77          & 18.13          & 34.66          & 50.48          & 0.637             & 0.633            \\
                                 & 10000            & $\times$           & \textbf{61.55} & \textbf{73.84} & \textbf{82.67} & \textbf{19.02} & \textbf{35.45} & \textbf{51.01} & \textbf{0.666}    & \textbf{0.665}   \\ \hdashline
\multirow{4}{*}{L64}             & 1024             & $\checkmark$       & 50.16          & 64.20          & 74.63          & 17.54          & 33.01          & 48.16          & 0.542             & 0.548            \\
                                 & 4096             & $\checkmark$       & 58.29          & 70.83          & 80.02          & 17.86          & 33.59          & 49.33          & 0.619             & 0.617            \\
                                 & 4096             & $\times$           & 59.72          & 72.22          & 81.38          & \textbf{18.54} & \textbf{34.77} & 50.50          & 0.644             & 0.649            \\
                                 & 10000            & $\times$           & \textbf{62.06} & \textbf{74.47} & \textbf{83.03} & 18.52          & 34.60          & \textbf{50.56} & \textbf{0.678}    & \textbf{0.681}   \\ \hdashline
\multirow{4}{*}{G128}            & 1024             & $\checkmark$       & 56.11          & 69.58          & 79.24          & 18.28          & 34.59          & 51.30          & 0.607             & 0.602            \\
                                 & 4096             & $\checkmark$       & 62.22          & 74.61          & 83.08          & 18.52          & 35.04          & 51.85          & 0.689             & 0.686            \\
                                 & 4096             & $\times$           & 62.94          & 75.33          & 84.08          & 18.88          & 35.99          & 52.52          & 0.694             & 0.692            \\
                                 & 10000            & $\times$           & \textbf{64.85} & \textbf{77.11} & \textbf{85.54} & \textbf{19.47} & \textbf{36.59} & \textbf{52.74} & \textbf{0.704}    & \textbf{0.708}   \\ \hdashline
\multirow{4}{*}{E128}            & 1024             & $\checkmark$       & 56.40          & 70.12          & 79.98          & 17.95          & 34.48          & 50.90          & 0.615             & 0.608            \\
                                 & 4096             & $\checkmark$       & 62.84          & 75.16          & 83.72          & 18.35          & 35.38          & 51.97          & 0.695             & 0.694            \\
                                 & 4096             & $\times$           & 64.07          & 76.45          & 84.76          & \textbf{19.49} & \textbf{36.61} & \textbf{53.25} & 0.700             & 0.697            \\
                                 & 10000            & $\times$           & \textbf{65.38} & \textbf{77.56} & \textbf{85.87} & 18.63          & 35.89          & 52.63          & \textbf{0.713}    & \textbf{0.714}   \\ \hdashline
\multirow{4}{*}{U128}            & 1024             & $\checkmark$       & 59.42          & 72.93          & 82.49          & 18.80          & 35.48          & 52.15          & 0.632             & 0.622            \\
                                 & 4096             & $\checkmark$       & 64.75          & 77.15          & 85.77          & \textbf{18.94} & 36.46          & 53.66          & 0.717             & 0.716            \\
                                 & 4096             & $\times$           & 66.31          & 78.62          & 86.83          & 18.62          & \textbf{36.50} & \textbf{53.89} & 0.713             & 0.709            \\
                                 & 10000            & $\times$           & \textbf{67.40} & \textbf{79.21} & \textbf{87.25} & 18.58          & \textbf{36.50} & 53.56          & \textbf{0.729}    & \textbf{0.732}   \\ \hline
\end{tabular}
}
\label{tab:discriminative-power}
\end{table*}

We evaluate the impact of kernel fusion on inference efficiency across all model variants using various keypoint counts. Throughput is measured for each architecture while processing between 1,024 and 16,384 keypoints. As illustrated in Fig.~\ref{fig:kernel_fusion}, where both axes use a $\log_2$ scale, the fused kernel implementation consistently delivers significantly higher throughput than the non-fused baseline. This efficiency gap widens as the model scale and the number of keypoints increase, demonstrating the superior scalability of our hardware-aware design.

The scaling behavior highlights the distinct advantages of our approach. Without kernel fusion, the FPS halves almost immediately as the number of sampling points doubles. This indicates that the system is bound by the high overhead of intermediate tensor management. In contrast, our fused strategy maintains stable performance across a much broader range of sampling densities. For larger configurations, a notable performance drop-off only occurs when the number of keypoints exceeds 10,000. This confirms that kernel fusion effectively minimizes memory bottlenecks and ensures real-time performance even under high-density conditions.

\begin{figure*}[!htbp]
    \centering
    \includegraphics[width=\textwidth]{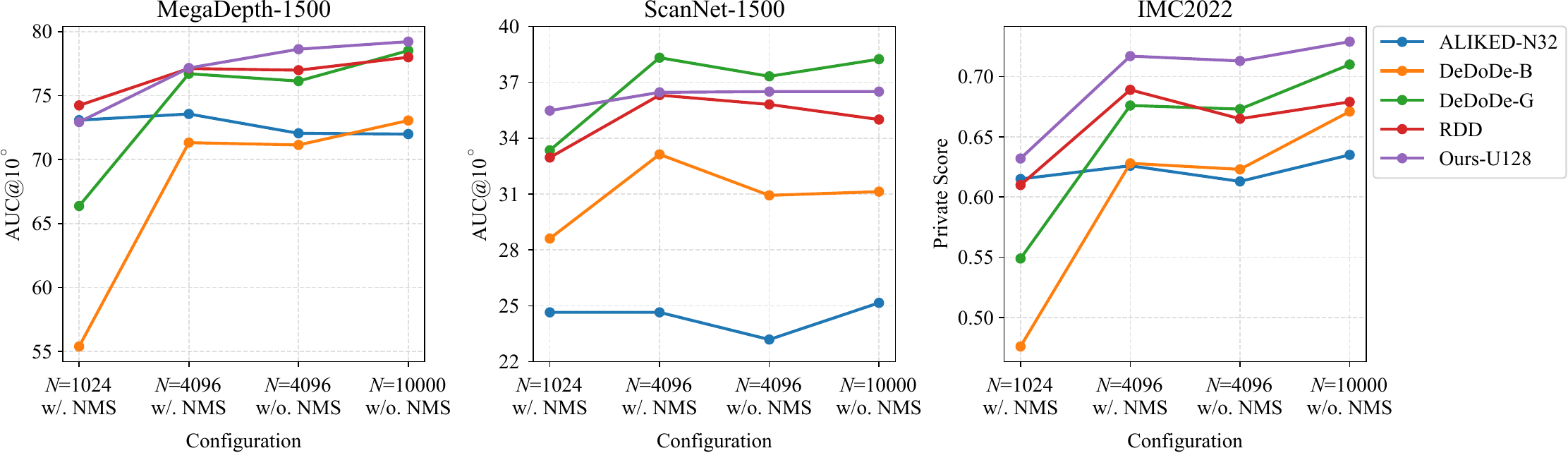}
    \caption{\raggedright \textbf{Performance under varying keypoint density on relative pose estimation.} Results compare U128 against ALIKED-N32, DeDoDe-B, DeDoDe-G, and RDD.}
    \label{fig:density_vs_relative_pose}
\end{figure*}

\subsubsection{Descriptor Discriminability Across Varying Spatial Densities}

Different downstream tasks demand different keypoint densities, yet descriptor matching performance varies with density as well. To validate how matching accuracy responds to density, we evaluate relative pose estimation under four configurations that span sparse to dense sampling: 1024 keypoints with NMS, 4096 keypoints with NMS, 4096 keypoints without NMS, and 10,000 keypoints without NMS.

Table~\ref{tab:discriminative-power} reports the results. Higher density consistently yields stronger matching accuracy across all evaluated models. The gains are particularly pronounced for smaller models. On MegaDepth-1500 and IMC2022, the configuration with 10,000 keypoints and no NMS achieves the highest scores for every model scale. On ScanNet-1500, the same configuration still produces strong results, though the margin over the 4096-without-NMS setting narrows and occasionally reverses. Indoor scenes with sparse texture offer fewer stable detection points than outdoors, which may limit performance gains.

We further compare U128 against ALIKED-N32, DeDoDe-B, DeDoDe-G, and RDD on relative pose estimation across density levels. Fig.~\ref{fig:density_vs_relative_pose} presents the results. At 1024 keypoints, most methods produce lower accuracy than at their higher-density settings. ALIKED holds steadier scores under this reduction, consistent with its training strategy built around sparse supervision, but its accuracy ceiling stays lower and it gains little from extra keypoints. DeDoDe exhibits a more noticeable decline when keypoints become sparse. RDD tracks closely at low point counts but does not gain from raising the limit. Across the full density range, our U128 model delivers leading performance and exhibits the most consistent upward trend as density increases.

\begin{figure*}[!htbp]
    \centering
    \includegraphics[width=\textwidth]{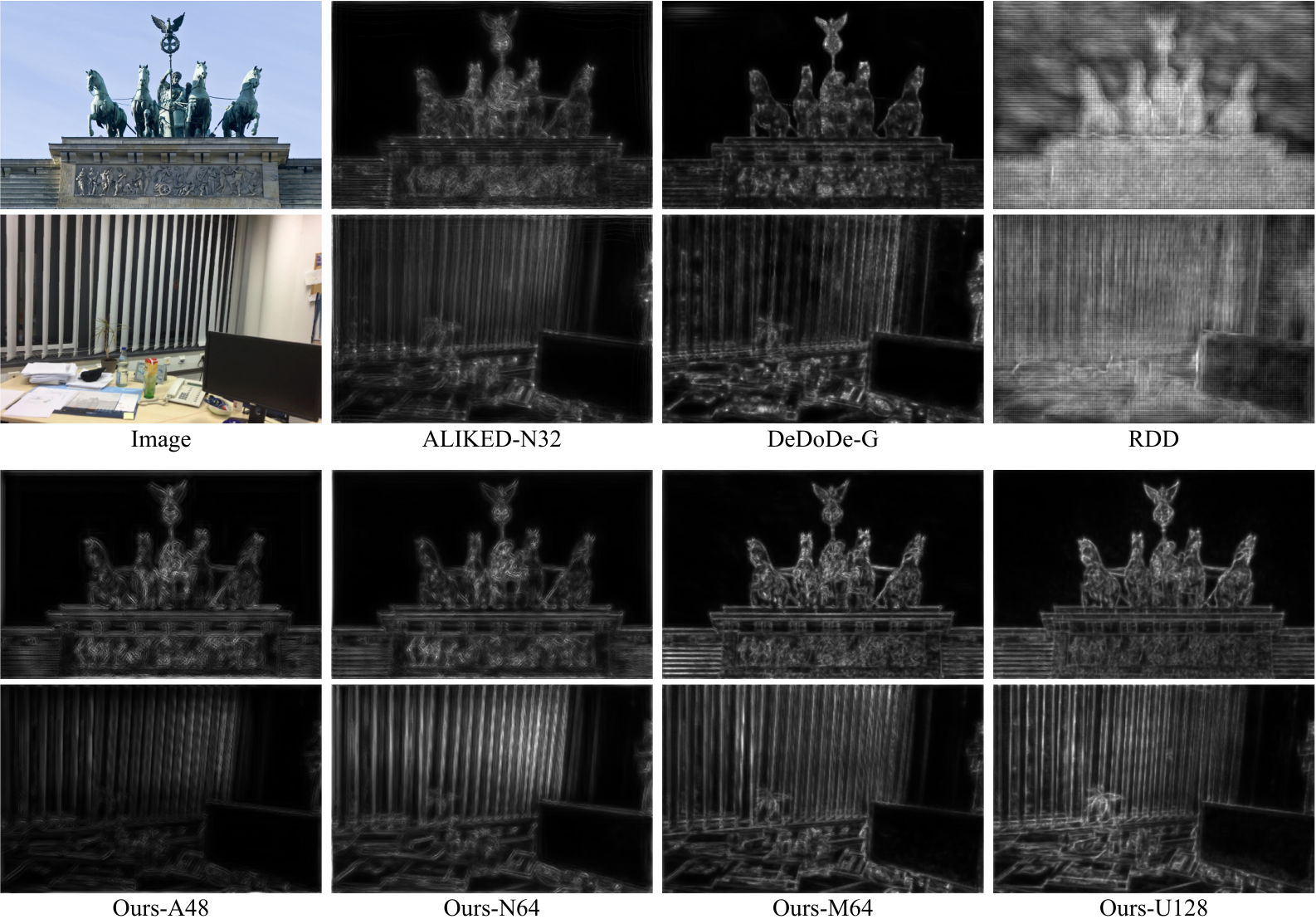}
    \caption{\raggedright \textbf{Spatial derivative visualization of dense descriptor maps for ALIKED, DeDoDe-G, RDD, and four CLIDD model variants.} The descriptor map spatial derivative reveals local correlation patterns. RDD exhibits blurred textures with grid-like artifacts from interpolation. CLIDD models produce sharp and uniform differential textures across all model scales.}
    \label{fig:cat_diff}
\end{figure*}

\begin{figure*}[!htbp]
    \centering
    \includegraphics[width=\textwidth]{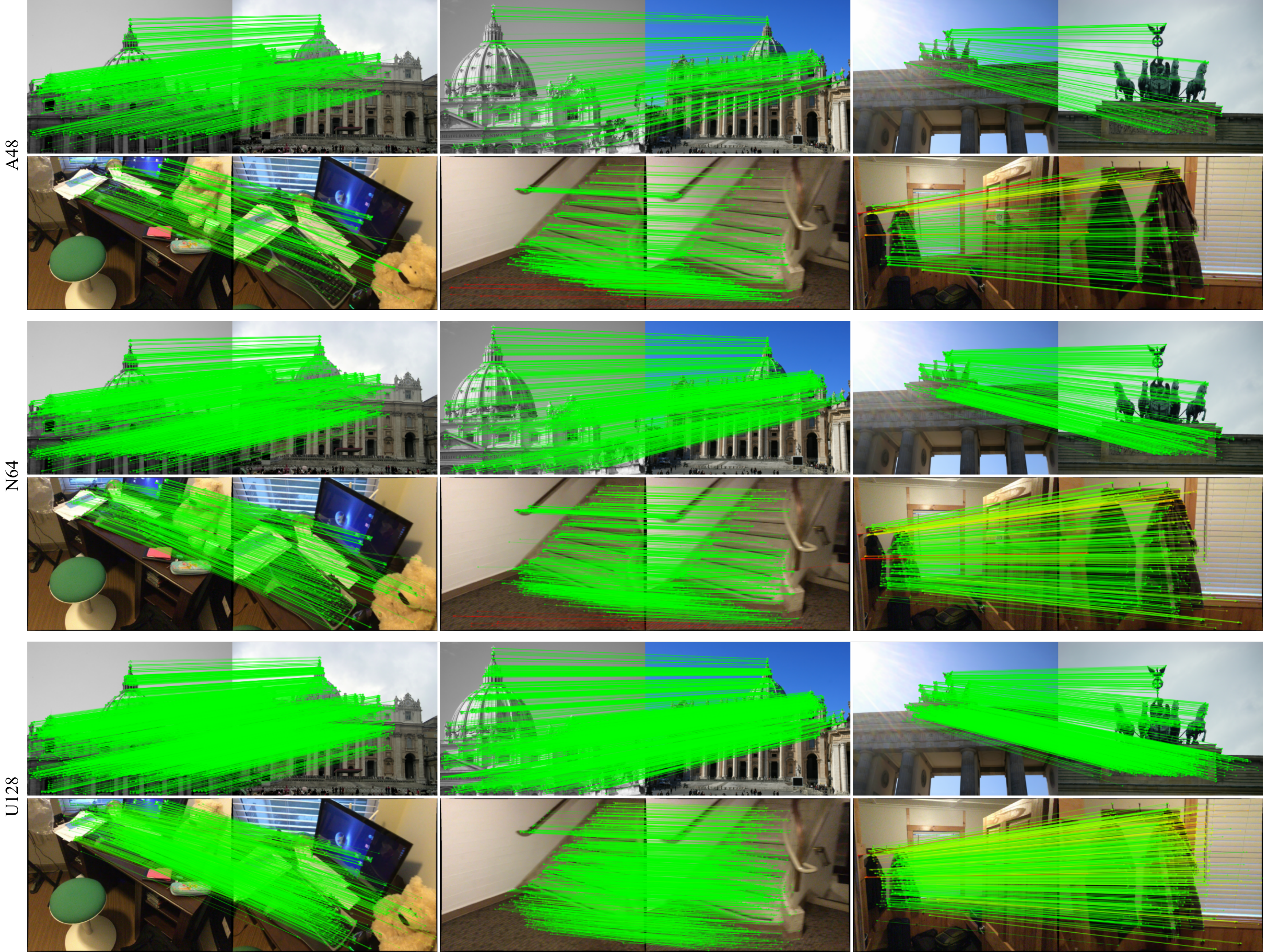}
    \caption{\raggedright \textbf{Dense keypoint matching results with 10,000 keypoints and without NMS.} We compare our lightweight models A48 and N64 against our largest U128.}
    \label{fig:extra_dense}
\end{figure*}

To verify that descriptor discriminability persists when NMS is removed and keypoints pack densely, we examine whether nearby descriptors remain distinguishable. We compute the spatial derivative of dense descriptor maps, producing a normalized visualization where sharper and more uniform textures indicate weaker local correlation. Fig.~\ref{fig:cat_diff} presents results for ALIKED, DeDoDe-G, RDD, and four CLIDD model variants. RDD relies on interpolated features from a low-resolution feature map, and its differential map reveals blurred textures with visible grid-like artifacts that reflect the smoothing introduced by interpolation. ALIKED and DeDoDe-G both produce sharper differential maps. However, ALIKED concentrates its strongest differential responses on sparse interest points as a consequence of its sparsely supervised training targets, leaving non-peak regions with weaker gradient. DeDoDe-G generates clean textures, though doing so requires maintaining full-resolution feature maps throughout the network. In contrast, all CLIDD models produce sharp and uniform differential textures, from the lightweight A48 variant to the high-capacity U128 model. This uniformity confirms that independent per-layer sampling directly from each feature layer's output resolution effectively breaks the local correlation, and the benefit holds across all model scales. Fig.~\ref{fig:extra_dense} provides a direct visual counterpart, showing matching results at 10,000 keypoints without NMS for three variants that span the full model range. All variants produce valid correspondences even at this extreme density, with matches growing visibly denser and cleaner as model capacity increases.

Collectively, these experiments demonstrate that CLIDD maintains strong descriptor discriminability across a wide range of spatial densities, from 1024 keypoints to 10,000 without NMS. This robustness to density variation makes the method a flexible foundation for tasks with differing demands on keypoint count and spatial distribution.

\subsubsection{Geometric Transformation Robustness}

To assess whether the deformable sampling points learned by CLIDD maintain consistency under different image transformations, we analyze their behavior under scale change, in-plane rotation, homographic warping, and perspective distortion using our U128 model. Fig.~\ref{fig:crosspage} presents the sampling offsets across the 1/2, 1/8, and 1/32 feature scales under each transformation. All sampling positions are projected to the original image resolution to facilitate direct comparison across transformations.

\begin{figure*}[!htbp]
  \centering
  \begin{subfigure}[b]{\textwidth}
    \centering
    \includegraphics[width=\textwidth]{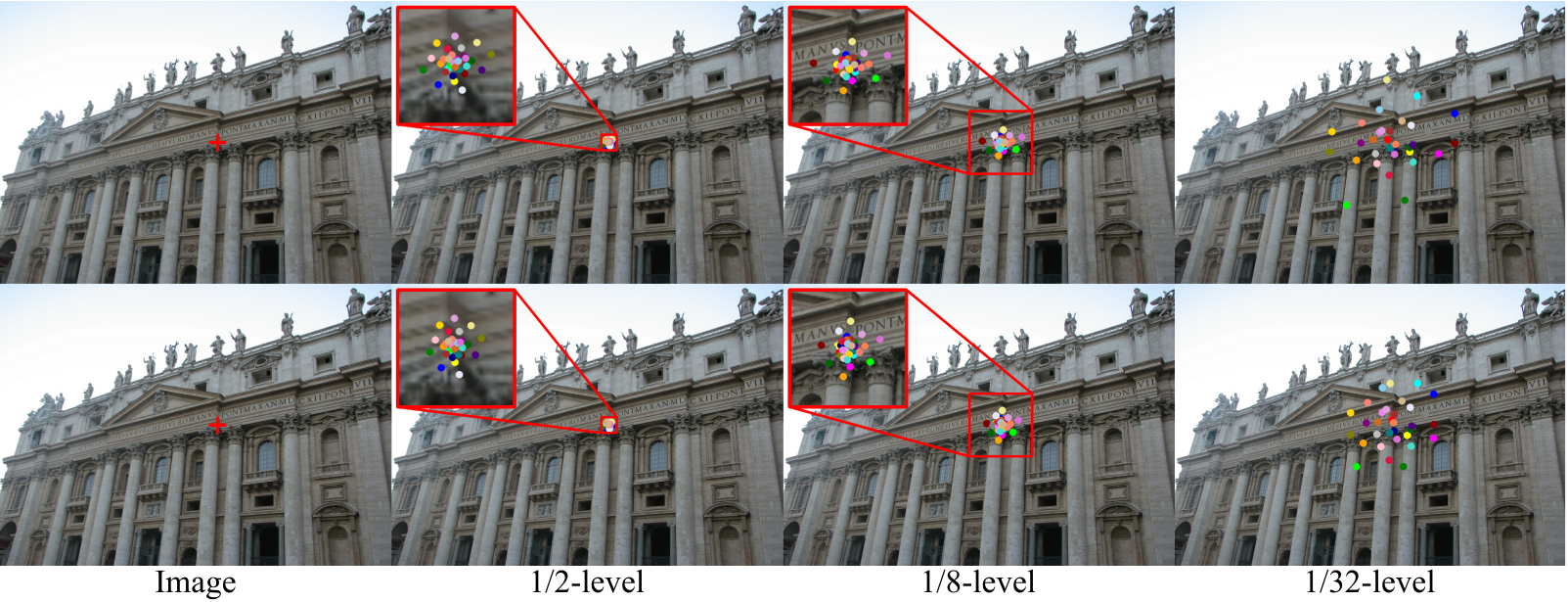}
    \captionsetup{skip=2pt}
    \subcaption{Scale}
  \end{subfigure}
  \vspace{2pt}
  
  \begin{subfigure}[b]{\textwidth}
    \centering
    \includegraphics[width=\textwidth]{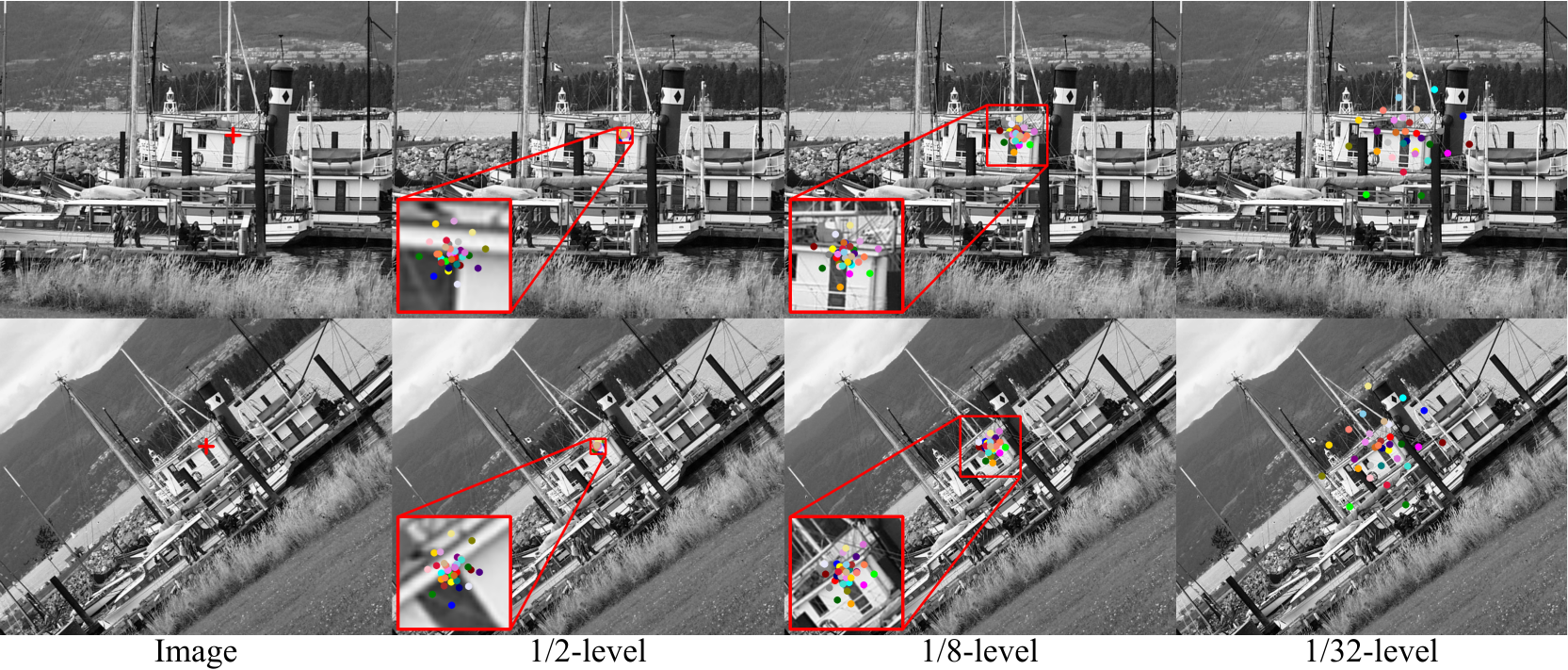}
    \captionsetup{skip=2pt}
    \subcaption{Rotation}
  \end{subfigure}
  \vspace{2pt}

  \begin{subfigure}[b]{\textwidth}
    \centering
    \includegraphics[width=\textwidth]{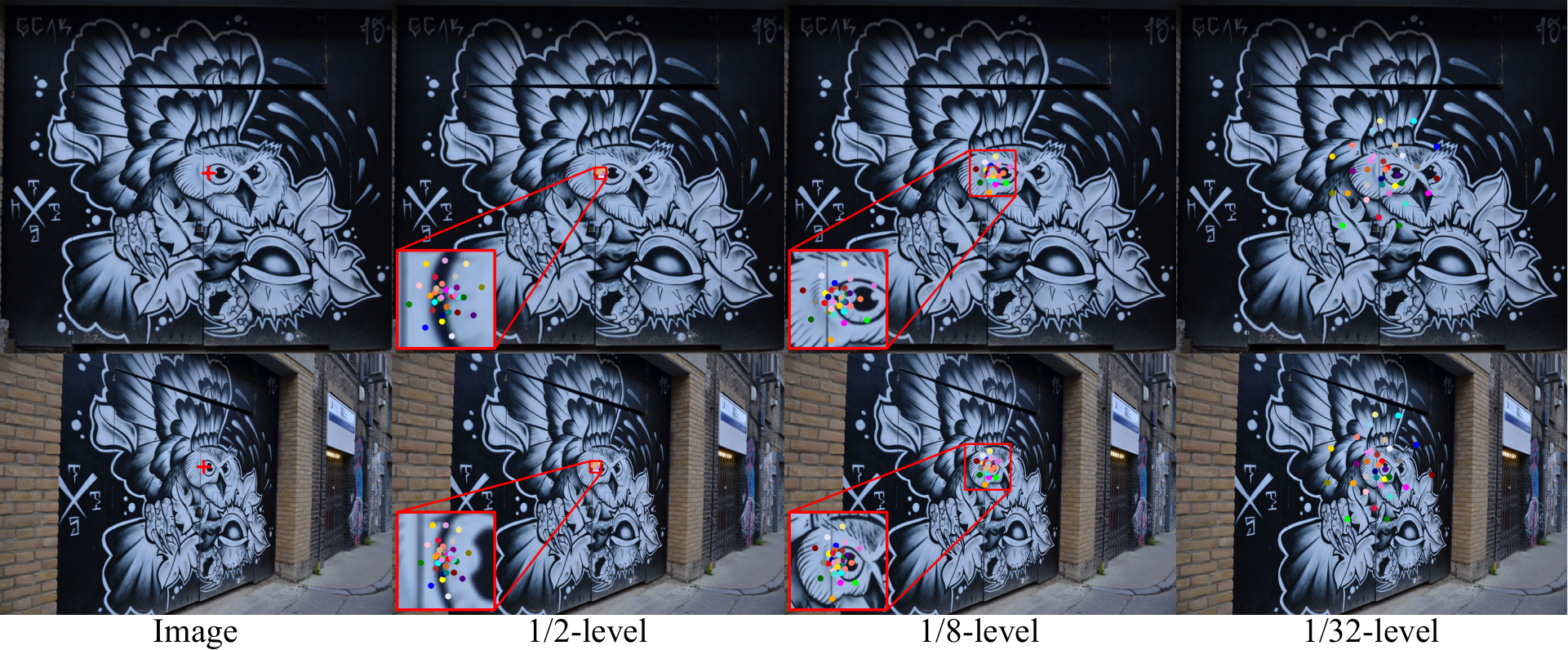}
    \captionsetup{skip=2pt}
    \subcaption{Homography}
  \end{subfigure}
  
  \captionsetup{labelformat=empty}   
  \caption{} 
  \captionsetup{labelformat=simple} 
\end{figure*}

\begin{figure*}[!ht]
  \ContinuedFloat     
  \setcounter{subfigure}{3}      
  \centering
  
  \begin{subfigure}[b]{\textwidth}
    \centering
    \includegraphics[width=\textwidth]{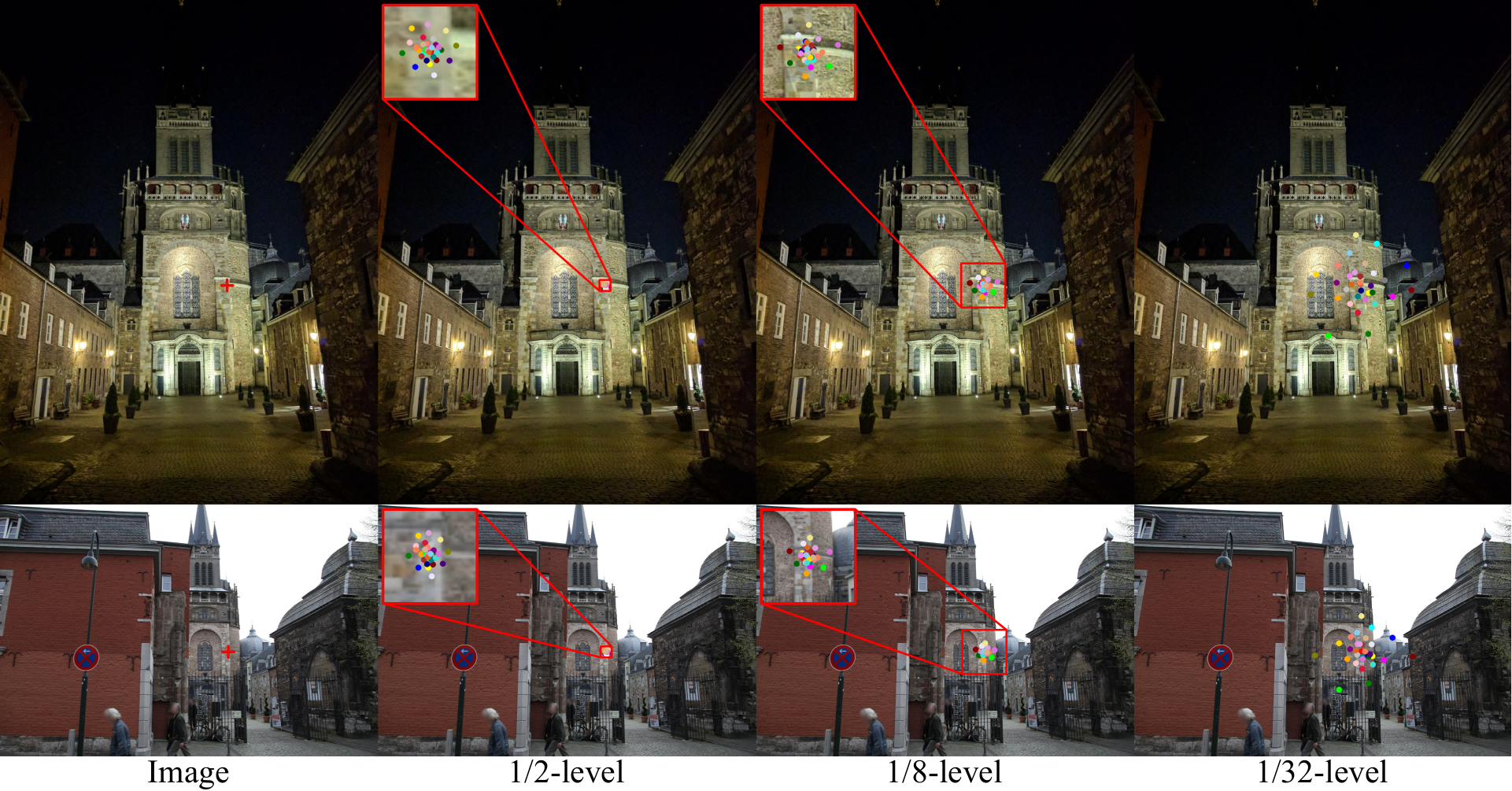}
    \captionsetup{skip=2pt}
    \subcaption{Perspective}
  \end{subfigure}

  \captionsetup{skip=3pt}
  \caption{\raggedright \textbf{Analysis of learned sampling offsets under scale, rotation, homography, and perspective transformations.} For the scale transformation, sampling offsets are computed from the original image and its 150\% enlarged version, with all points visualized at a common scale to enable direct comparison. All sampling points are displayed at the original image resolution.}
  \label{fig:crosspage}
\end{figure*}

A clear layer-wise pattern emerges from the spatial distribution of sampling points across feature scales. At the 1/2 scale, offsets remain compact and tightly localized around each keypoint. At the 1/8 scale, the sampling range expands to a visibly broader region. At the 1/32 scale, the offsets extend across a substantially wider spatial extent, reaching distant positions far from the keypoint. This scale-dependent distribution confirms that each feature scale contributes a qualitatively distinct form of information to the final descriptor. Shallow layers contribute localized structural features, while deep layers supply broader contextual information. Thus, CLIDD can preserve the distinct role of each feature scale even when the input undergoes geometric variation.

A closer examination across transformation types reveals how this robustness manifests under each form of geometric variation. Under scale change, sampling offsets at the 1/2 and 1/8 scales remain largely stable, while some at the 1/32 scale exhibit noticeable displacement at positions farther from the keypoint. This pattern has two causes. First, resizing forces predicted offsets to span a wider spatial range. At shallow layers, the offset range is naturally compact and remains well within the receptive field after resizing. At deeper layers, the receptive field is fixed while the image grows, so the widened offset range risks extending past the region of reliable correspondence. Second, deformable sampling applies an explicit bound on offset magnitudes to prevent excessive drift. This bound is inherited from deformable convolutions and naturally limits how well far-range sampling can adapt to scale changes. Under in-plane rotation, the 1/2-scale offsets exhibit a rotation that closely tracks the image transformation, while the alignment fidelity progressively weakens at the 1/8 and 1/32 scales. This gradation follows from the nature of standard convolutions, which offer no built-in rotation equivariance. Without explicit rotation augmentation during training, deeper layers progressively lose the capacity to track rotational alignment. Homographic warping follows a consistent trend, with shallower features preserving greater local invariance and deeper features extending to wider contextual regions. Under perspective changes, the 1/2 and 1/8 scales continue to yield precise local estimates while the 1/32 scale proves more effective at reaching distant sampling points. The distinct role of each scale therefore persists rather than collapsing under severe geometric distortion.

An important qualification is that descriptor extraction operates without access to the image against which matching will later be performed. Therefore, the network cannot compute exact offsets for any specific transformation. From a probabilistic perspective, the network learns which sampling configurations tend to remain stable across the distribution of geometric variations encountered during training. The patterns described above reflect this learned bias rather than a guarantee of deterministic equivariance.

These observations confirm that CLIDD produces robust offsets across common geometric transformations. Each feature scale develops a specialized role that persists under geometric variation, and the layered structure of offset distributions remains intact across all tested conditions. This consistency provides a direct explanation for why the method maintains reliable matching performance even under challenging viewing geometries.

\section{Conclusions}

In this work, we presented a systematic framework for efficient and high-precision local feature representation based on Cross-Layer Independent Deformable Description. CLIDD generates sparse descriptors directly from independent multi-scale layers. Through the Cross-Layer Predictor and the Layer-Independent Sampler, our approach preserves the distinct characteristics of each feature hierarchy, enabling specialized offset patterns to strengthen descriptor discriminability. It also bypasses the construction of a unified dense feature map, eliminating the associated memory and computation overhead. A hardware-aware kernel fusion strategy further integrates sampling and aggregation into a single operation, enabling real-time inference on resource-constrained edge devices. We demonstrated that our scalable architecture family spans a wide spectrum of model variants optimized for diverse deployment constraints, with all variants trained through a composite protocol of metric learning and knowledge distillation. Experimental results across multiple geometric benchmarks demonstrate leading matching performance together with state-of-the-art inference throughput across the scalable model family. This work confirms that high structural fidelity and discriminative power can be achieved without the heavy computational burden of high-resolution feature maps or dense fusion modules.

While our framework achieves strong descriptor performance, the current detection training protocol relies on distilling keypoint outputs from existing methods. In future work, we plan to leverage the trained descriptors themselves to identify high-quality keypoints, enabling self-supervised detection training that reduces reliance on external distillation.

\section*{Data Availability}

All the data used in this paper come from open-source datasets that are freely available for anyone to download. The demo and weights of our work are available at \url{https://github.com/HITCSC/CLIDD}.

\section*{Acknowledgments}

This work was supported by the National Natural Science Foundation of China (No. 62403063), and Lejoin Intelligence (Shenzhen) Co., Ltd.

\bibliographystyle{spbasic}      
\bibliography{ref}   

\end{document}